\documentclass[a4paper, 11pt]{article}

\usepackage[american]{babel}
\usepackage[utf8]{inputenc}
\usepackage[T1]{fontenc}
\usepackage{lmodern}
\usepackage{csquotes}
\usepackage{fullpage}
\usepackage{authblk}
\usepackage{amsthm}
\usepackage{amssymb}
\usepackage{mathtools}

\usepackage{pdflscape} % for landscape orientation
\usepackage{tikz}
\usepackage[colorinlistoftodos,
            prependcaption,
            textsize=small]{todonotes}
\usepackage{microtype}

\usepackage[colorlinks,
            citecolor=blue,
            urlcolor=blue,
            linkcolor=blue]{hyperref}
\vfuzz \hfuzz
\providecommand{\keywords}[1]{\noindent\textbf{Keywords:} #1}

\usetikzlibrary{external,arrows.meta,positioning}

\tikzsetexternalprefix{figures/}

\tikzexternaldisable

\newtheoremstyle{plain} % redefinition of "plain" style
{1em}% Space above
{1em}% Space below
{}% Body font
{}% Indent amount 1
{\bfseries}% Theorem head font
{.}% Punctuation after theorem head
{.5em}% Space after theorem head 2
{}% Theorem head spec (can be left empty, meaning ‘normal’)

\newtheorem{theorem}{Theorem}%[section]
\newtheorem{assumption}[theorem]{Assumption}

\newtheorem{definition}[theorem]{Definition}
\newtheorem{example}[theorem]{Example}
\newtheorem{corollary}[theorem]{Corollary}
\newtheorem{proposition}[theorem]{Proposition}

\usepackage{natbib}

\newcommand{\FF}{\Lambda} 

 \usepackage{subcaption}
 \usepackage{enumitem}
 \usepackage{tabularx}
 \usepackage{booktabs}
 \usepackage{tikz}
\usetikzlibrary{calc}
 \usepackage{xcolor}
 \usepackage{stmaryrd} % iverson bracket
\usepackage[noend]{algpseudocode}
\usepackage{algorithm}

\newlength{\algofontsize}
\begin{document}

\title{Decision-focused predictions via pessimistic
bilevel optimization: complexity and algorithms}

\author[1]{V\'ictor Bucarey L\'opez\thanks{Corresponding author.}}
\author[2]{Sophia Calder\'on}
\author[3]{Gonzalo Muñoz}
\author[4]{Frédéric Semet}
\affil[1,2]{Institute of Engineering Sciences, Universidad de O'Higgins, Chile, \textit{victor.bucarey@uoh.cl}}
\affil[3]{ Department of Industrial Engineering, Universidad de Chile, Chile, \textit{gonzalo.m@uchile.cl}}
\affil[4]{Inria Lille-Nord Europe, France, \textit{frederic.semet@centralelille.fr}}

\maketitle
% \listoffigures

\begin{abstract}
Dealing with uncertainty in optimization parameters is an important and longstanding challenge. Typically, uncertain parameters are predicted accurately, and then a deterministic optimization problem is solved. However, the decisions produced by this so-called {\it predict-then-optimize} procedure can be highly sensitive to uncertain parameters. In this work, we contribute to recent efforts in producing {\it decision-focused} predictions, i.e., to build predictive models that are constructed with the goal of minimizing a {\it regret} measure on the decisions taken with them.
We begin by formulating the exact expected regret minimization as a pessimistic bilevel optimization model. Then, we show computational complexity results of this problem, including its membership in NP. In combination with a known NP-hardness result, this establishes NP-completeness and discards its hardness in higher complexity classes.
Using duality arguments, we reformulate it as a non-convex quadratic optimization problem. Finally, leveraging the quadratic reformulation, we show various computational techniques to achieve empirical tractability. We report extensive computational results on shortest-path and bipartite matching instances with uncertain cost vectors.
Our results indicate that our
approach can improve training performance over the approach of Elmachtoub and Grigas (2022), a state-of-the-art method for decision-focused learning. \\
\end{abstract}

%%%%%%%
% Text 
%%%%%%

\section{Introduction}\label{sec:Intro}

Decision-making processes often involve uncertainty in input parameters, which is an
important and longstanding challenge. Commonly, a two-stage approach is employed: firstly, training a machine learning (ML) model to estimate the uncertain input accurately, and secondly, using this estimate to tackle the decision task. This decision task is typically an optimization problem. Classical machine learning methods focus mainly on minimizing prediction errors on the parameters, disregarding the impact these errors might have on the subsequent optimization task.  This approach overlooks how inaccurate predictions can negatively influence the optimization solution, potentially leading to decisions of poor quality.

In recent years, {\it decision-focused learning} (DFL)  (also known as smart predict-then-optimize (SPO)) approaches, in which prediction and optimization tasks are integrated into the learning process, have received significant attention. In this approach, a machine learning model is specifically trained to enhance the effectiveness of the whole decision-making process. This involves combining, during the training phase, the prediction and the optimization in a single model. 

In this work, we focus on linear optimization problems where the coefficients of the cost vector $c \in \mathbb{R}^n$ are unknown, but we have at hand a vector of correlated features $x$. Our goal is to train a parametric machine learning model $m(\omega, x)$, where $\omega$ is the vector of parameters of the machine learning model, so that the impact of the prediction error on the whole decision process is minimal. The typical measure of how a prediction performs in the decision process is the {\it regret}: the excess of cost in the optimization task caused by prediction errors.

As we will see in the following sections, finding the model $m(\omega, x)$ that minimizes the regret can be formulated as a pessimistic bilevel optimization problem. In fact, authors in \cite{elmachtoub2022smart} define the unambiguous-SPO loss function as a pessimistic regret: among all the solutions that minimize the predicted objective function, it penalizes the one that hurts the most when it is evaluated with the true cost vector. 

Unfortunately, the complexity of finding these models and scalability are two major roadblocks to this DFL approach. Consequently, the literature has mainly focused on stochastic gradient-based approaches via approximations of the loss function through a surrogate convex loss function and/or solving a relaxed optimization problem (see \cite{mandi2023decision}).
Here, we follow a different path, take a step back, and focus on carefully studying the mathematical object behind the {\it exact} expected regret minimization. Our main hypothesis is that by understanding the mathematical object and designing new methods for solving the pessimistic bilevel optimization problem, better predictions and better algorithms can be developed. 

 The contributions of this work are the following: (i) we formulate the expected regret minimization problem as a pessimistic bilevel optimization problem; (ii) we prove that the problem belongs to the NP complexity class, which settles NP-completeness and makes it unlikely for it to be higher in the polynomial hierarchy (i.e., the problem is not $\Sigma_2^P$-hard, unless the hierarchy collapses); (iii) we show that, under mild assumptions, determining if the regret is 0 is polynomial-time solvable; (iv) we reformulate the bilevel pessimistic formulation as a non-convex quadratically-constrained quadratic program (QCQP), which can be tackled by current optimization technology for moderate sizes; (v) we propose heuristics to improve the solution procedure based on the quadratic reformulation; (vi) we conduct a comprehensive computational study on shortest path and bipartite matching instances. An early version of this work was published as a short paper in the conference proceedings of CPAIOR 2024 \citep{bucarey2024decision}, {where the single-level reformulation and a local-search heuristic were presented. We extend this work by providing theoretical results regarding the complexity of this problem and an alternate descent direction method. We finally extend the experiments by including the bipartite matching problem in the computational study.}

We note that, in parallel to the preparation of this extended version, the work of \cite{jimenez2025pessimistic} also proposed to use a pessimistic bilevel optimization approach for DFL. The method proposed is considerably different, as \cite{jimenez2025pessimistic} proposes a column-and-constraint generation and a branch-and-cut approach. A close comparison is subject of future work.

\subsection{Problem setting}  We consider a nominal optimization problem with a linear objective function:

\begin{equation} 
P(c):  \quad   z^*(c) := \min_{v  \in V} c^\top v \label{prob:setup}
\end{equation}
In this work, we restrict $V$ to be a non-empty polytope, i.e. a non-empty bounded polyhedron. For a given $c$, we define $V^*(c)$ as the set of optimal solutions to \eqref{prob:setup}.

 In our setting, the value of $c\in \mathbb{R}^n$ is not known, but we have access to a dataset $\mathcal D = \{(x^i, c^i)\}_{i=1}^N$ with historical observations of $c$ and correlated feature vectors $x\in \mathbb{R}^K$. 
Given these observations, and for a fixed set of parameters $\omega$, we can empirically measure the sensitivity of the decisions given by the predictions using an average {\it regret}:

\begin{align} \label{eq:mregret}
   \mbox{Regret}(\mathcal{D}, \omega) := \,  \max_{v} \quad & \frac{1}{N}\sum_{i\in [N]} \left(c^i {}^\top v^i - z^*(c^i) \right) , \\
     \mbox{s.t.}\quad  & v^i \in V^*( m(\omega ,x^i) ) \quad \forall i \in [N]
\end{align}

Here, we are comparing the {\it true} optimal value $z^*(c^i)$ with the value $c^i {}^\top v^i$, which is the ``true cost'' of a solution that is optimal for the prediction $m(\omega,x^i)$.
To find the values of $\omega$ that minimize the regret \eqref{eq:mregret}, we must solve the following pessimistic bilevel optimization problem.

\begin{align} \label{expected_regret_sample}
    \min_\omega \max_{v^i \in V^*( m(\omega ,x^i)) } \quad \frac{1}{N} \sum_{i\in [N]} \left( c^i{}^\top v^i - z^*(c^i) \right)
\end{align}

In this formulation, there are three optimization problems involved: i. the lower-level problem optimizing $m(\omega ,x^i)^\top v$ over $V$; ii. over all the possible optimal solutions of the latter, take the one with the maximum (worst) regret. This corresponds to the pessimistic version of the bilevel formulation, defining the pessimistic regret. iii. Minimize the pessimistic regret using $\omega$ as a variable.
In what follows, we use the notation $\hat{c}^i(\omega) := m(\omega, x^i)$.

\subsection{Importance of the pessimistic approach} \label{sec:example}

To illustrate the relevance of the \emph{pessimistic} approach (in contrast to the \emph{optimistic} one), we consider the following linear optimization problem with two variables:
\begin{equation*}\label{example}
    P(c): \quad \min_v \quad  c_1 v_1 + c_2 v_2 
    \quad \text{s.t.} \quad v_1 + v_2 \le 1, \quad
    v_1, v_2 \ge 0.
\end{equation*}
Suppose we observe the following three instances of the true cost vectors, each associated with a single feature value:

\begin{equation*}
  (x^1,c^1) = \left(0, \begin{pmatrix} -3 , -2 \end{pmatrix} \right)^\top, \,
  (x^2,c^2) = \left(1, \begin{pmatrix} -2,-5 \end{pmatrix} \right), \,
  (x^3,c^3) = \left(2, \begin{pmatrix} -2, 0 \end{pmatrix} \right)^\top.
\end{equation*}

Our goal is to estimate a linear regression model of the form $m(\omega, x)_j = \hat{c}_j(\omega) = \omega_{0j} + \omega_{1j} x, \quad \text{for } j = 1, 2.$

We begin by analyzing the optimistic version, which corresponds to dropping the outer maximization $\max_{v \in V^*(\hat{c}^i(\omega))}$ in \eqref{expected_regret_sample}. This is equivalent to setting $\omega_{0j} = \omega_{1j} = 0$ for both $j = 1,2$, since in that case $\hat{c} = (0,0)$ and the solution set becomes $V^*((0,0)) = V$ (i.e., every feasible $v \in V$ is optimal).

Thus, the optimistic model ``guesses'' the correct decision in all cases, resulting in zero regret. However, the \textbf{pessimistic} model evaluates the regret over all possible solutions in $V^*((0,0))$ and selects the worst-case one---in this case, $v_1 = v_2 = 0$. The regrets under this decision are 3, 5, and 2 for observations 1, 2, and 3, respectively, yielding an average regret of $\frac{10}{3}$. Thus, this ``optimistic regret'' does not adequately measure the sensitivity of the decision with respect to the predictive model.

In contrast, a classical linear regression performs better under the pessimistic regret criterion. The minimum squared error solution for this regression is
\[
\omega = \begin{pmatrix}
    -2.83 & 0.50 \\
    -3.33 & 0.99
\end{pmatrix},
\]
which induces the correct decision for observation 3 and results in an average regret of $\frac{4}{3}$.

Solving the bilevel pessimistic formulation exactly leads to an even lower regret of $\frac{1}{3}$, with an optimal regression model
\[
\omega = \begin{pmatrix}
    -1 & -1 \\
    -4 & 1
\end{pmatrix}.
\]
This solution correctly predicts the optimal decision for observations 2 and 3 (those with the highest regret) and only fails on observation 1.

We also evaluate the SPO+ estimator proposed by \cite{elmachtoub2022smart}, computed via a linear programming formulation. This method yields an average regret of 1.

\begin{figure}[htpb]
	\centering
	\includegraphics[width=0.9\linewidth]{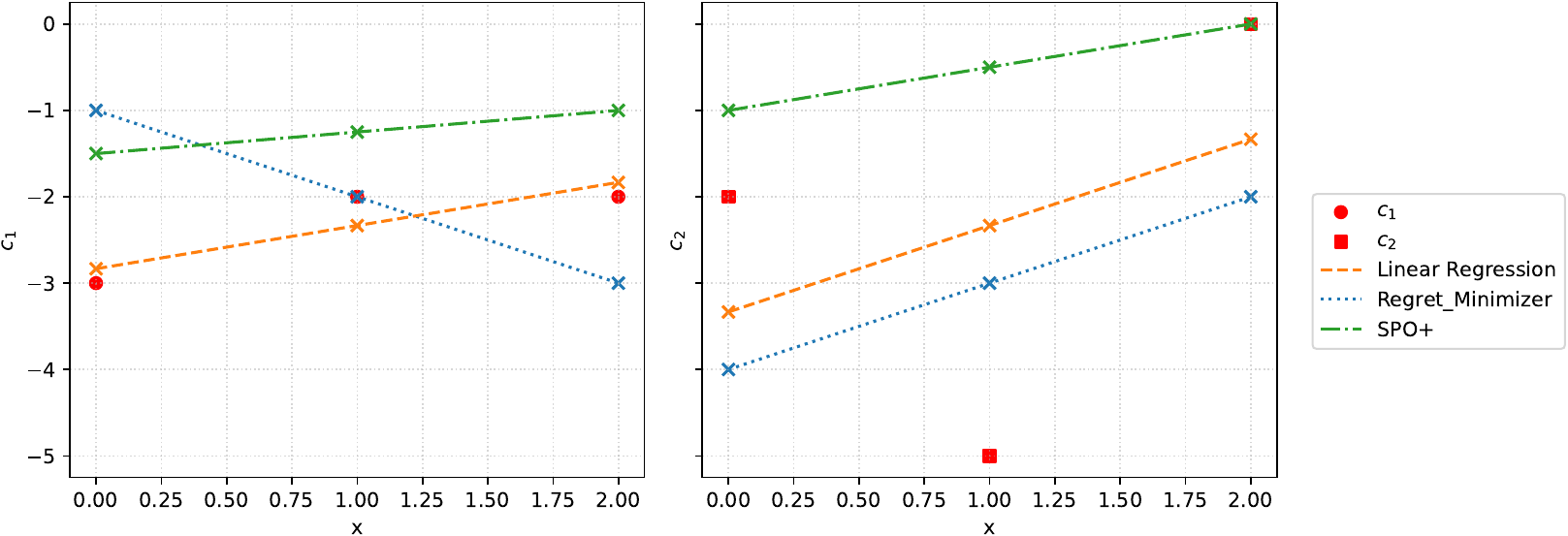}
	\caption{Linear regression, SPO+ estimator from \cite{elmachtoub2022smart} and an exact regret minimizer for the numerical example of Section \ref{sec:example}. Red circles and squares represent the input data. Dash-dotted lines correspond to the different solutions for the linear model $m(\omega, x)$.}
	\label{fig:example}
\end{figure}

Figure~\ref{fig:example} summarizes the solutions and outcomes for each approach.

\subsection{Literature Review}

\paragraph{Bilevel optimization.}  

Bilevel optimization problems are hierarchical ones: they consist of an optimization problem (called upper-level or leader problem)
that contains in its constraints optimality conditions of other problems, called lower-level or follower problems \cite{dempe2020bilevel}. In the presence of several optimal solutions at the lower-level problems, the way that the solution is chosen opens two approaches: the optimistic approach, also known as the strong solution, in which the solution chosen is the one that favors the upper-level optimization problem; and the pessimistic, also known as the weak solution, that chooses the one that worsens the most the upper-level objective. 

 Optimistic bilevel optimization is NP-hard even when the upper and lower-level problems are linear. Even having compact/efficient formulations can be a difficult task \cite{kleinert2020there}. See \cite{kleinert2021survey} for a survey on mixed-integer programming techniques in bilevel optimization.  
The necessity of establishing the pessimistic solution for bilevel optimization problems was first raised by \cite{leitmann1978generalized} and studied in several articles (e.g., \cite{loridan1996weak,aboussoror2005weak}).
Until recently, it was a common belief that the pessimistic approach is much more difficult than the optimistic one.
However, it has been established that a pessimistic \emph{linear} problem can be transformed equivalently into an optimistic one. \cite{henke2025coupling,zeng2020practical}
We refer to readers to \cite{wiesemann2013pessimistic,liu2018pessimistic,liu2020methods} for comprehensive surveys of theoretical background and methods for bilevel pessimistic optimization problems.
 We note that the hardness results in bilevel optimization rule out the existence of {\it efficient general-purpose} techniques that could be applied in our setting.
 In addition, the equivalence between pessimistic and optimistic was established for the linear case; as we will see, our model is non-linear.

\paragraph{Decision-focused learning.} 
As mentioned earlier, and to the best of our knowledge, the existing methods developed for training decision-focused predictions are primarily based on estimating the gradient of how predictions impact a specific loss function. The main challenge in these approaches is to estimate the changes in the optimal solution with respect to the model parameters, known as optimization mapping. According to the recent survey by \cite{mandi2023decision}, these methods can be classified into four categories: i. those that compute the gradient of the optimization mappings analytically, as in \cite{amos2017optnet}; or ii. solving a smooth version of the optimization mapping, as seen in \cite{wilder2019melding,mandi2020interior,ferber2020mipaal}; or  iii. smoothing the optimization mapping by random perturbations \cite{PogancicPMMR20, niepert2021implicit} ; or iv. those solving a surrogate loss function that approximates regret, as discussed in \cite{elmachtoub2022smart,mulamba2021contrastive,mandi2022decision}. 

Our study diverges from these approaches. Instead, to find regret-minimizing models, we leverage the pessimistic bilevel optimization formulation \eqref{expected_regret_sample}, and by employing duality arguments, we formulate it as a single-level non-convex quadratic model. A related method for minimizing the expected regret was proposed by \cite{jeong2022exact}: they cast the problem, not explicitly, as an optimistic bilevel optimization problem and use a symbolic reduction to solve this problem. However, as mentioned in \cite{elmachtoub2022smart} and discussed in Section \ref{sec:example}, casting this problem as an optimistic bilevel optimization problem may lead to undesirable predictions.
 The recent work of \cite{jimenez2025pessimistic} also proposes an \emph{exact} pessimistic bilevel optimization approach.
 
In terms of the complexity of \eqref{expected_regret_sample}, the results of \cite{elmachtoub2022smart} imply that this problem is NP-hard. They show that minimizing the regret function generalizes the minimization of the 0-1 loss function, which is known to be NP-hard \citep{ben2003difficulty}.
Here, we complement this by showing membership in NP of a decision version of \eqref{expected_regret_sample}.

\section{Complexity results} \label{s:complexity}

In this section, we focus on studying the complexity of \eqref{expected_regret_sample} in a basic setting: when the predictive model $m(\omega,x)$ is linear. Since we are predicting cost {\it vectors}, we can assume our set of parameters $\omega$ are in matrix form, i.e., $m(\omega,x)=\omega x$.

As mentioned in the previous section, the results of \cite{elmachtoub2022smart} imply that \eqref{expected_regret_sample} is NP-hard since it generalizes the minimization of the 0-1 loss function, which is known to be NP-hard \citep{ben2003difficulty}. 
In this section, we begin by showing membership in NP, thus establishing NP-completeness of the problem.
Note that Buchheim  \cite{buchheim2023bilevel} recently showed that linear bilevel optimization, both optimistic and pessimistic, belongs to NP. However, this result is not directly applicable to \eqref{expected_regret_sample} since the leader and follower variables interact non-linearly in the follower's problem, even when $m(\omega, x)$ is a linear function.

\subsection{Membership in NP}
Let us consider the following decision version of \eqref{expected_regret_sample}.

\begin{definition}
The decision problem $\mbox{SIMPLE-REGRET}$ is defined as follows. Given $(c^i, x^i)_{i=1}^N$ a collection of $N$ rational vectors and matrices, a polytope $V$, and a rational number $M$ decide if there exists $\omega$ such that
\begin{align} \label{regret_simpler}
\max_{v^i \in V^*(\omega x^i  )} \quad \frac{1}{N} \sum_{i\in [N]}  (c^i{}^\top v^i - z^*(c^i)) \quad  \leq \quad M.
\end{align}
\end{definition}
\noindent We note that \eqref{regret_simpler} is obtained from \eqref{expected_regret_sample} by restricting $m(\omega, x)= \omega x$.

\begin{theorem}\label{theo:membershipNP}
 $\mbox{SIMPLE-REGRET} \in \mbox{NP}$
\end{theorem}
\begin{proof}
It suffices to show that, for any ``Yes'' instance of $\mbox{SIMPLE-REGRET}$, there is a polynomially-sized $\omega$ such that \eqref{regret_simpler} holds.
Indeed, if we have such an $\omega$, we can simply compute the optimal faces of each of the $N$ lower-level problems (which can be done in polynomial time), and then solve the resulting maximization problem.
Let us show that such an $\omega$ exists.\\

Consider an arbitrary instance of $\mbox{SIMPLE-REGRET}$, with $V=\{v\,:\, Av \geq b\}$, and suppose there exists $ \hat \omega$ such that the value of
\begin{align*} \label{regret_NP}
\max \quad & \frac{1}{N}\sum_{i=1}^N (c^i{}^\top v^i - z^*(c^i))\quad \text{s.t} \quad  v^i\in \arg\min \,\big \{ (\hat{\omega} x^i )^\top v \,:\, A v \geq b \big\}
\end{align*}

\noindent is less or equal than $M$. We will show we can modify $\hat \omega$ to have polynomial size and not change any of the argmins. Since we assume $V$ is always non-empty and bounded (which can be verified in polynomial time), strong duality always holds. 
The dual of the $i$-th lower-level problem reads
\begin{align*}
    \max\quad & (\rho^i)^\top b \quad
    \text{s.t}\quad  (\rho^i)^\top A = \hat \omega  x^i, \quad
 \rho^i \geq 0.
\end{align*}
For each $i$, let us consider $\hat{\rho}^i$ in the relative interior of the optimal face of the dual. This means that the optimal face of the $i$-th primal can be described as
\[F^i(\hat \omega) := \{v \in V\,:\, a_j^\top v = b_j,\, j\,:\, \hat{\rho}^i_j > 0\}. \]
We claim that any $(\tilde \omega, \tilde \rho)$ satisfying
\begin{subequations}\label{eq:system}
\begin{align}
    (\tilde{\rho}^i)^\top A & = \tilde \omega x^i  && \forall i\in [N]\\
    \tilde{\rho}^i_j & \geq  1 && \hat{\rho}^i_j > 0,\, \forall i\in [N]\\
    \tilde\rho^i_j &= 0 && \hat{\rho}^i_j = 0,\, \forall i\in [N]
\end{align}
\end{subequations}
is such that $\arg\min\{ (\tilde{\omega} x^i )^\top v \,:\,  A v \geq b\} = F^i(\hat \omega)$ for every $i\in[N]$.
Indeed, if there are such $\tilde \omega$ and $\tilde \rho$, then
\begin{align*}
 \arg\min\{ (\tilde{\omega} x^i )^\top v \,:\,  A v \geq b\} 
    = &\arg\min\{ (\tilde{\rho}^i)^\top A v \,:\,  A v \geq b\}
    \\
    = &\arg\min\left\{ \sum_{j:\hat{\rho}^i_j > 0} \tilde{\rho}^i_j a_j^\top v \,:\,  A v \geq b\right\}
\end{align*}

The last objective function is lower bounded by $\sum_{j:\hat{\rho}^i_j > 0} \tilde{\rho}^i_j b_j$. Moreover, this lower bound is met if and only if $a_j^\top v = b_j,  \, \forall j\,:\, \hat{\rho}^i_j > 0.$
Therefore, we obtain that each $\tilde \rho^i$ is dual optimal for $\min\{ (\tilde{\omega} x^i )^\top v \,:\,  A v \geq b\}$ and $\arg\min\{ (\tilde{\omega} x^i )^\top v \,:\,  A v \geq b\} = F^i(\hat \omega)$.\\

To conclude, note that \eqref{eq:system} is always feasible, as a rescaled version of $(\hat \omega, \hat \rho)$ satisfies the system. Since the coefficients of \eqref{eq:system} are given by the entries of $A, x^i$, zeros, and ones, we conclude that there must be a $(\tilde \omega, \tilde \rho)$ of {\it polynomial size} that satisfies the system. 
\end{proof}
\begin{corollary}
$\mbox{SIMPLE-REGRET}$ is $\mbox{NP}$-complete.
\end{corollary}
\begin{corollary}
$\mbox{SIMPLE-REGRET}$ is not $\Sigma_2^P$-hard, unless the polynomial hierarchy collapses at the second level.
\end{corollary}
As a final remark on this subsection, we note that the previous proof strategy can be used to show two facts in the optimization context of \eqref{expected_regret_sample} when $m(\omega, x)= \omega x$.

\begin{corollary}\label{cor:attainable}
    Consider \eqref{expected_regret_sample} when $m(\omega, x)= \omega x$. The regret function \eqref{eq:mregret} only has a finite number of values, and, furthermore, the minimum regret \eqref{expected_regret_sample} is always attained.
\end{corollary}
The piecewise constant nature of the regret \eqref{eq:mregret} is known in the literature (e.g., \cite{poganvcic2019differentiation,demirovic2020dynamic}). The fact that it only attains a finite number of values in our setting, although perhaps expected, is not entirely direct.
Also, recall that some bilevel optimization problems do not attain their optimal values.

The proof of Corollary \ref{cor:attainable} follows from the fact that, once we fix the optimal faces of each lower-level problem, the regret is fixed. For completeness, we provide the proof of Corollary \ref{cor:attainable} in Appendix A.

\subsection{Polynomial-time solvable cases}

In this subsection, we show that in many cases, determining if the regret is 0 can be done in polynomial time.
This is in line with the polynomial solvability of checking if zero loss can be achievable in empirical risk minimization in multiple cases. For example, checking if zero loss can be achieved in 0-1 loss minimization (i.e., if the data is separable) can be solved in polynomial time: it amounts to finding a separating hyperplane among the two classes.
However, we note that, perhaps counterintuitively, having zero regret is \emph{not} equivalent to determining if there exists $\omega$ such that 
\[\omega x^i = c^i \qquad \forall i \in [N].\]

We illustrate this in the following example.

\begin{example}\label{example:zeroregret}
Consider the polytope $V = [0,1]^2$ and the following two observations
\[(x^1,c^1) = \left(1, \begin{pmatrix} -1, -2 \end{pmatrix}^\top, \right)\,
  (x^2,c^2) = \left(-1, \begin{pmatrix} 1 , 1 \end{pmatrix}^\top \right).\] It is not hard to see that 
\[\arg\min\{c^1{}^\top v\,:\, v\in [0,1]^2\} = \{(1,1)^\top \}\quad \text{and} \quad \arg\min\{c^2{}^\top v\,:\, v\in [0,1]^2\} = \{(0,0)^\top \}\]

If we take, for example, $\omega = (-1,-1)^\top$, we have that
\[\arg\min\{(\omega x^1)^\top v\,:\, v\in [0,1]^2\} = \{(1,1)^\top \} \text{ and } \arg\min\{ (\omega x^2)^\top v\,:\, v\in [0,1]^2\} = \{(0,0)^\top \}\]

And from this, we can deduce that $\omega$ yields zero regret. However, it is not hard to see that there is no $\omega$ such that $\omega x^i = c^i, \, i=1,2,$ as $(\omega x^1, \omega x^2)$ are always collinear in this example, but $(c^1,c^2)$ are linearly independent.
\end{example}

Our polynomial-time solvability result uses the following assumption.

\begin{assumption}\label{assumption:uniqueopt}
    The input data $(c^i, x^i)_{i=1}^N$ and the polytope $V$ are such that $\arg\min\{c^i{}^\top v\,:\, v\in V\}$ is a singleton, for all $i\in [N]$.
\end{assumption}

Assumption \ref{assumption:uniqueopt} may seem restrictive, but it can be expected for real data to satisfy it; if the $c^i$ are drawn from a non-atomic distribution, for example, Assumption \ref{assumption:uniqueopt} is satisfied with probability 1.
Additionally, note that this does not imply that the follower will have a unique solution in general: it can still be that the optimal face for $\omega x^i$ over $V$ is not a singleton.
Finally, note that Assumption \ref{assumption:uniqueopt} can be checked in polynomial time.

We are now ready to describe a polynomial time solvable case for \eqref{expected_regret_sample}.

\begin{theorem}\label{thm:polytime}
If the input for $\mbox{SIMPLE-REGRET}$ is restricted to $M=0$ (i.e., determining if there is a solution with zero regret) and the data $(c^i, x^i)_{i=1}^N$ and the polytope $V$ satisfy Assumption \ref{assumption:uniqueopt}, then the problem can be solved in polynomial time.
\end{theorem}

We provide the proof of Theorem \ref{thm:polytime} in Appendix B. As a last comment in this subsection, we conjecture that, without Assumption \ref{assumption:uniqueopt}, the problem of determining if there is a solution with zero regret becomes NP-hard.
This is somewhat counterintuitive since some NP-hard loss minimization problems (as 0-1 loss minimization) become easy when the question is whether zero loss is achievable or not.
However, we conjecture this based on the discussion on Example \ref{example:zeroregret}: one can have zero regret with data that cannot be fit perfectly in the traditional sense, and this can be obtained with a non-trivial ``alignment" of the optimal faces $V^*(\omega x^i)$.

\section{A non-convex quadratic reformulation}
\label{sec:nonconvex}

While NP-hardness of \eqref{expected_regret_sample} makes it unlikely for us to find a worst-case efficient method for solving it, we can still hope to find methods with good practical performance.
In this section, we will apply duality arguments, supported by the assumption that $V$ in problem \eqref{prob:setup} is non-empty and bounded, in order to obtain a more manageable formulation of the problem.

As before, we assume \eqref{prob:setup} has the following form:

\begin{subequations}\label{shortestpathLP2}
\begin{align}
 \min_{v} \quad & c^\top v\quad \text{s.t.} \quad  A v\geq b \label{shortestpathLP2b} 
\end{align}
\end{subequations}

\noindent Our predicted costs have the form $m(\omega,x)$ for some feature vector $x$ and parameters $\omega$ (to be determined), and the terms $z^*(c^i)$ are constant. Thus, an equivalent formulation of our (pessimistic) regret-minimization problem is:
\begin{equation}\label{eq:bilevelpessimistic}
\begin{aligned}
   \min_{\omega} \max_{v} & \quad \frac{1}{N} \sum_{i \in [N]} (c^{i})^\top v^{i}\\
   \text{s.t.} & \quad v^{i} \in \arg\min_{\tilde v^i} \quad  m(\omega,x^i)^\top \tilde{v}^i\\
    & \quad \quad \quad \quad \text{s.t.} \quad \quad A\tilde{v}^i\geq b 
\end{aligned}
\end{equation}

\subsection{Duality arguments}

{A common approach to solving optimistic bilevel problems involving convex lower-level problems is to reformulate them by replacing the lower-level problems with their optimality (Karush-Kuhn-Tucker or KKT) conditions (see \cite{kleinert2021survey}). In this subsection, we follow the same type of argument twice to achieve a single-level reformulation of the pessimistic bilevel problem \eqref{eq:bilevelpessimistic}.}

Since the feasible region of the lower-level problem in \eqref{eq:bilevelpessimistic} is a non-empty polytope, which is unaffected by $\omega$, we can apply LP duality and reformulate \eqref{eq:bilevelpessimistic} as:
\begin{subequations}\label{eq:bilevelpessimistic2}
\begin{align}
   \min_{\omega} \max_{v,\rho, \alpha} & \quad \frac{1}{N} \sum_{i \in [N]} (c^{i})^\top v^{i} \label{objective}\\
   \text{s.t.} & \quad  A v^i\geq b && \forall i\in [N] \label{primal1}\\
   & \quad A^\top \rho^i  = m(\omega,x^i) && \forall i\in [N] \label{dual1}\\
    & \quad \rho^i  \geq 0 && \forall i\in [N] \label{dual2} \\
     & \quad m(\omega,x^i)^\top v^i \leq b^\top \rho^i && \forall i\in [N] \label{strongduality}
\end{align}
\end{subequations}
In this formulation \eqref{primal1} imposes primal feasibility, \eqref{dual1}-\eqref{dual2} impose dual feasibility, and \eqref{strongduality} imposes strong duality. We note that strong duality is typically written as an equality constraint; however, the $\geq$ inequality always holds due to weak duality.
We purposely opted for imposing strong duality directly instead of complementary slackness, as the latter yields non-linear inequalities on $(v,\rho)$.
\\

The inner maximization problem of \eqref{eq:bilevelpessimistic2} is an LP, which is feasible for every value of $\omega$. This is true because it represents a primal-dual system of a linear problem over a non-empty polytope, meaning it always has a solution. Moreover, since the objective function in \eqref{objective} involves only the $v$ variables, which are bounded, we know that strong duality holds, allowing us to take the dual once again. This yields the following reformulation of \eqref{eq:bilevelpessimistic}:
\begin{subequations}\label{eq:bilevelpessimistic-final}
\begin{align}
    \min_{\omega, \mu, \delta, \gamma} \, & \sum_{i\in [N]} \left( b^\top \mu^i + m(\omega,x^i)^\top \delta^i
   \right)\\
   \text{s.t. }  & A^\top \mu^i + m(\omega,x^i) \gamma^i  = \frac{1}{N} c^i && \forall i\in [N]\\
    & A\delta^i - b \gamma^i \geq 0 && \forall i\in [N] \\
   & \mu^i \leq 0,\, \gamma^i \geq 0  && \forall i\in [N]
\end{align}
\end{subequations}
This is a single-level, non-convex quadratic problem.

\subsection{Shortest path as a bounded linear program}

In this work, and motivated by \cite{elmachtoub2022smart}, we consider the shortest path problem with unknown cost vectors. It is well known that this problem can be formulated as a linear program using a totally unimodular constraint matrix. However, the feasible region may not be bounded, as the underlying graph may have negative cycles, which correspond to extreme rays of the corresponding polyhedron.

To apply our framework (which relies on duality arguments), we need to assume no negative cycles exist for every possible prediction $m(\omega, x)$. For this reason, we make the following assumption.
\begin{assumption}\label{assumption:acyclic}
The underlying graph $G$ defining the shortest path is directed acyclic.
\end{assumption}

Under this assumption, we can safely formulate the shortest path problem as \eqref{shortestpathLP2}, and every extreme point solution will be a binary vector indicating the shortest path.
Additionally, the inner maximization problem in \eqref{eq:bilevelpessimistic2} always has a binary optimal solution, as its feasible region is an extended formulation of the optimal face of the lower-level problem.
From this discussion, we can guarantee that under Assumption \ref{assumption:acyclic}, formulation \eqref{eq:bilevelpessimistic-final} is valid for the shortest path problem with uncertain costs.

In our experiments, we also consider weighted bipartite instances. These instances can be directly formulated as a linear program with a bounded and integral feasible region, and thus, we do not need any extra assumption on them.

\section{Solution methods}

As mentioned earlier, in this work, we focus on the case when $m(\omega, x)$ is a linear model. Given its excellent performance, we will use the SPO+ loss function as a baseline. 

\subsection{SPO+} Authors in \cite{elmachtoub2022smart} define the SPO+ loss function as a convex surrogate loss, which serves as an upper bound on the true regret. The authors derive an informative subgradient for this loss, which can be utilized in any subgradient descent-based approach. Additionally, they present a linear programming formulation for cases where the nominal problem is linear, and the model for estimation is also linear. This surrogate loss is defined as
\begin{align*}
 \ell_{SPO}(c, m(\omega, x)) =& \max_{v \in V} \left\{ (c-2m(\omega, x))^\top v \right\} + 2m(\omega, x)^\top v^*(c) - z^*(c).
\end{align*}
By using \eqref{shortestpathLP2} as a nominal problem, the problem that minimizes the expected SPO+ loss can be cast as the following LP:
\begin{subequations} \label{SPO+_lp}
    \begin{align}
        \min_{\omega, \rho} \quad\frac{1}{N}  &\sum_{i\in [N]} \left( -b^\top \rho_i  + 2m(\omega, x^i)^\top v^*(c^i) \right) \\%- z^*(c)\\
        \mbox{s.t. }& \,  -\rho_i^\top A = c^i - 2m(\omega, x^i) & i \in [N]\\
        &\rho^i\ge 0 & i \in [N].
    \end{align}
\end{subequations}
\noindent Variables $\rho$ are dual variables associated to constraints $Av \ge b$ of the nominal problem (\ref{shortestpathLP2}). We consider this formulation as a starting point and as a benchmark against which we will compare our methods.

\subsection{Local Search}
The intermediate reformulation presented in \eqref{eq:bilevelpessimistic2} can be seen as an unrestricted optimization problem of the form
\[
\min_\omega \Lambda(\omega).
\]
Here, $\Lambda(\omega)$ is a function that for each $\omega$ returns the optimal value of the inner maximization problem in  \eqref{eq:bilevelpessimistic2}. As mentioned above, for each $\omega$, $\Lambda(\omega)$ is a feasible linear problem. We propose the following simple local-search-based heuristic: given an initial incumbent solution $\omega_0$, we randomly generate $T$ new solutions in a neighborhood of $\omega_0$, evaluate the regret for each, and update the incumbent solution with the one with the smallest regret. We repeat these steps during $L$ iterations. The procedure is detailed in Algorithm \ref{algorithm1}.

\begin{algorithm}[t]
\caption{Local-search based algorithm}
\label{algorithm1}
\begin{algorithmic}[1]
\State {\bf Input} Training data,  Starting model parameters $\omega_0$. 
\State {\bf Hyperparameters}: size of neighbourhood $\epsilon$, sample size $T$, maximum number of iterations $L$. 
\State Solve $\FF(\omega_0)$
\For {$i = 0,\ldots,N$}
\State Sample $T$ parameters in the neighbourhood of $\omega_i$: $\omega_t \gets \omega_i + \epsilon \cdot \mathcal{N}(0,1)$
\State Compute  $\FF(\omega_t) \quad \forall t \in \{1, \ldots, T \}$ 
\State Update $\omega_{i+1} \gets \mbox{arg}\min_{t=1,...,T} F(\omega_t)$
\EndFor
\end{algorithmic}
\end{algorithm}

\subsection{Penalization} 
Based on the ideas of \cite{aboussoror2005weak}, we propose the following related formulation: we fix the variables $\gamma^i$ in \eqref{eq:bilevelpessimistic-final} to have all the same fixed value $\kappa$. This parameter $\kappa$ is set before optimization and can be seen as a hyperparameter of the optimization problem.
This results in the following model.
\begin{subequations}\label{eq:penalized-final}
\begin{align}
   \min_{\omega,\mu, \delta} \quad & \sum_{i\in [N]} \left( b^\top \mu^i + m(\omega,x^i)^\top \delta^i
   \right)\\
   \text{s.t.} \quad & A^\top \mu^i + m(\omega,x^i) \kappa = \frac{1}{N} c^i && \forall i\in [N]\\
    & A\delta^i - b \kappa \geq 0 && \forall i\in [N] \\
   & \mu^i \leq 0  && \forall i\in [N]
   \end{align}
\end{subequations}
This formulation corresponds to a slice of the formulation \eqref{eq:bilevelpessimistic-final} which removes all non-linearities of the constraints. Consequently, by adopting this approach, we solve an optimization problem with a quadratic non-convex objective and linear constraints.

We remark that this approach results in a problem that is not a traditional penalization (which typically yields relaxations) but rather a restriction of the problem. The name ``penalization'', which is used in \cite{aboussoror2005weak}, comes from a derivation of \eqref{eq:penalized-final}, which follows a similar approach to the one described in Section \ref{sec:nonconvex}. The difference lies in penalizing \eqref{strongduality} using $\kappa$ before taking the dual a second time.

\subsection{Alternating direction method}

We note that the non-convexities of formulation \eqref{eq:bilevelpessimistic-final} come from products between the model parameter $\omega$ and dual variables $\gamma$ or $\delta$.
Hence, if we fix either the $\omega$ variables or the variables $\gamma$ and $\delta$, we obtain linear programming problems. 
Specifically, if we fix $\omega$ to $\bar \omega$ the resulting LP reads
\begin{subequations}\label{eq:alternating_method_LP1}
\begin{align}
    \min_{\mu, \delta, \gamma} \, & \sum_{i\in [N]} \left( b^\top \mu^i + m(\bar\omega,x^i)^\top \delta^i
   \right)\\
   \text{s.t. }  & A^\top \mu^i + m(\bar\omega,x^i) \gamma^i = \frac{1}{N} c^i & \forall i\in [N]\\
    & A\delta^i - b \gamma^i \geq 0 & \forall i\in [N] \\
   & \mu^i \leq 0  & \forall i\in [N]\\
   & \gamma^i \geq 0 & \forall i\in [N].
\end{align}
\end{subequations}

Analogously, fixing $\bar{\gamma}$ and $\bar{ \delta}$ yields the LP
\begin{subequations}\label{eq:alternating_method_LP2}
\begin{align}
    \min_{\mu, \omega} \quad & \sum_{i\in [N]} \left( b^\top \mu^i +  m(\omega,x^i)^\top \bar{ \delta}^i
   \right)\\
   \text{s.t. }  & A^\top \mu^i + m(\omega,x^i) \bar{\gamma}^i = \frac{1}{N} c^i & \forall i\in [N]\\
       & \mu^i \leq 0   & \forall i\in [N]
\end{align}
\end{subequations}

Based on these observations, we propose Algorithm \ref{algorithm2} as a heuristic to find high-quality values for $\omega$.

\begin{algorithm}[htpb]
\caption{Alternating descent algorithm}
\label{algorithm2}
\begin{algorithmic}[1]
\State {\bf Input} Training data,  Starting model parameters $\omega_0$. 
\State {\bf Hyperparameters}: Maximum number of iterations $L$. 
\For {$i = 0,\ldots,L$}
\State Solve problem \eqref{eq:alternating_method_LP1} using $\bar{\omega}_{i}$. Retrieve optimal variables $\bar \delta_i$ and $\bar \gamma_i$
\State Solve problem \eqref{eq:alternating_method_LP2} using $\bar \delta_i$ and $\bar \gamma_i$ and retrieve a new vector of parameters $\bar{\omega}_{i+1}$ 
\EndFor
\State Return $\bar{\omega}_{L}$
\end{algorithmic}
\end{algorithm}

The following proposition follows directly from the definition of Algorithm \ref{algorithm2}.
\begin{proposition} \label{prop:alternating}
    The sequence $ \Lambda(\bar{\omega}^0), \Lambda(\bar{\omega}^1),  \ldots, \Lambda(\bar{\omega}^L) $ produced by Algorithm \ref{algorithm2} is non-increasing.
\end{proposition}

\subsection{Regression bounds and valid inequalities}

To prevent the solver from generating solutions with large coefficients in the non-convex QCQPs, and since the lower-level optimization problem is invariant to scalings of the objective, we can impose arbitrary bounds on the values of $\omega$. Any bound is valid, but we avoided small numbers to prevent numerical instabilities.\\

Additionally, to improve the performance of the optimization solver, we included the following valid inequality to \eqref{eq:bilevelpessimistic-final}:
\[ \sum_{i \in [N]} \left( b^\top \mu^i + m(\omega,x^i)^\top \delta^i \right)
    \geq \frac{1}{N}\sum_{i \in [N]} z^*(c^i)\]
This is a dual cut-off constraint. Its left hand side takes the same value as $\frac{1}{N} \sum_{i \in N} (c^{i})^\top v^{i}$ in \eqref{eq:bilevelpessimistic}, and thus, by optimality of $z^*$, the inequality holds.
This simple inequality provided considerable improvements in the solver's performance.

\section{Computational experiments}

\subsection{Computational set-up}

\paragraph{Data generation.}
We consider an adaptation of the data generation process described in \cite{elmachtoub2022smart} and \cite{tang2022pyepo}. The training data consists of $\{ (x^i,  c^i)\}_{i=1}^N$ generated synthetically the following way.

We consider two families of instances: small instances including values of $N = \{50, 100, 200\}$ where all methods can be run in moderate running times, and instances with $N = 1000$ to test the scalability of our approach.
Each dataset is separated into 70\% for training and 30\% for testing. Feature vectors are generated by sampling them from a standard normal distribution (mean zero and standard deviation equal to 1). We generate cost vectors by first generating the parameters $\omega$ of the model --representing the true underlying model-- and then using the following formula (see \cite{elmachtoub2022smart} and \cite{tang2022pyepo}):
\begin{equation}
    c^i_{a} = \bigg [ \frac{1}{3.5^{\text{Deg}}}\bigg(\frac{1}{\sqrt{K}} \big(\sum_{k=1}^K\omega_k x^i_{ak}  \big) +3  \bigg)^{\text{Deg}  } +1 \bigg] \cdot \varepsilon
\end{equation}
where $c^i_{a}$ is the component of the cost vector $c^i$ corresponding to the arc $a$ in the graph. 
The {\it Deg} parameter specifies the extent of model misspecification; as a linear model is used as a predictive model in the experiments, the higher the value of {\it Deg}, the more the relation between the features and cost coefficients deviates from a linear one (and thus, the larger the errors will be). Finally, a multiplicative noise term $\varepsilon$ is sampled randomly from a uniform distribution in $[0.5, 1.5]$. 
We perform our experiments by considering the values of the parameter Deg in $\{2, 8, 16\}$.

We consider two nominal problems: shortest path over a directed acyclic grid graph of 5 $\times$ 5 nodes; a maximum weight matching on a bipartite graph of 13 and 12 nodes.
To obtain graphs of similar sizes in both families of instances, we fix the number of edges of the bipartite matching graph to 40. 

We remark that the exact reformulation requires solving a challenging non-convex problem, and that the underlying problem is NP-complete; these are the main reasons why we focus on moderate instance sizes in this work. To provide some perspective, the reformulation involves $N\cdot \left(\#\mbox{Nodes} + 2\cdot \#\mbox{Edges} + 1\right) + K\cdot \#\mbox{Edges}$ variables and $N\cdot (\#\mbox{Nodes} + \#\mbox{Edges})$ constraints, from which $N\cdot \#\mbox{Edges}$ are quadratic non-convex constraints.

\paragraph{Algorithms.}

In our experiments, we consider the following sequence of steps to generate decision-focused predictions:
\begin{enumerate}
    \item Generate an initial solution using the SPO+ method computed by the linear program \eqref{SPO+_lp} (SPO)
    \item Improve the previous solution using Algorithm \ref{algorithm1} (LS)
    \item Either:
    \begin{enumerate}
        \item Solve \eqref{eq:penalized-final} (Penalized) using Gurobi with the solution produced by LS as a warm start
        \item Solve \eqref{eq:bilevelpessimistic-final} (Exact) using Gurobi with the solution produced by LS as a warm start
        \item Use the Alternating Method (Algorithm 2) with the solution produced by LS as the starting point
        \item Use the Alternating Method (Algorithm 2) with SPO as starting point
    \end{enumerate}
\end{enumerate}
In terms of computational efficiency, Step 1 involves solving one LP; Step 2 solves a fixed number of LPs; Steps 3(a) and 3(b) are non-convex quadratic problems; and Steps 3(c) and 3(d) are sequences of LPs.

This generates six different combinations that we test below, with their names indicating the sequence: SPO, SPO-LS, SPO-LS-EXA, SPO-LS-PEN, SPO-LS-ALT, and SPO-ALT. 

In all variants, we limit the entire suite of algorithms to one hour in the following way. We set a maximum time limit of 20 minutes for the SPO-LS part, leaving the remaining time for executing either EXA, PEN, or ALT. In our results, we provide the intermediary performance of SPO-LS. We remark that SPO can be solved in seconds. For instance, in our experiments, a typical instance with $N=1000$ is solved in less than a minute.\\

\paragraph{Hyperparameters.}

\begin{figure}[t]
	\centering
	\includegraphics[width=0.95\linewidth]{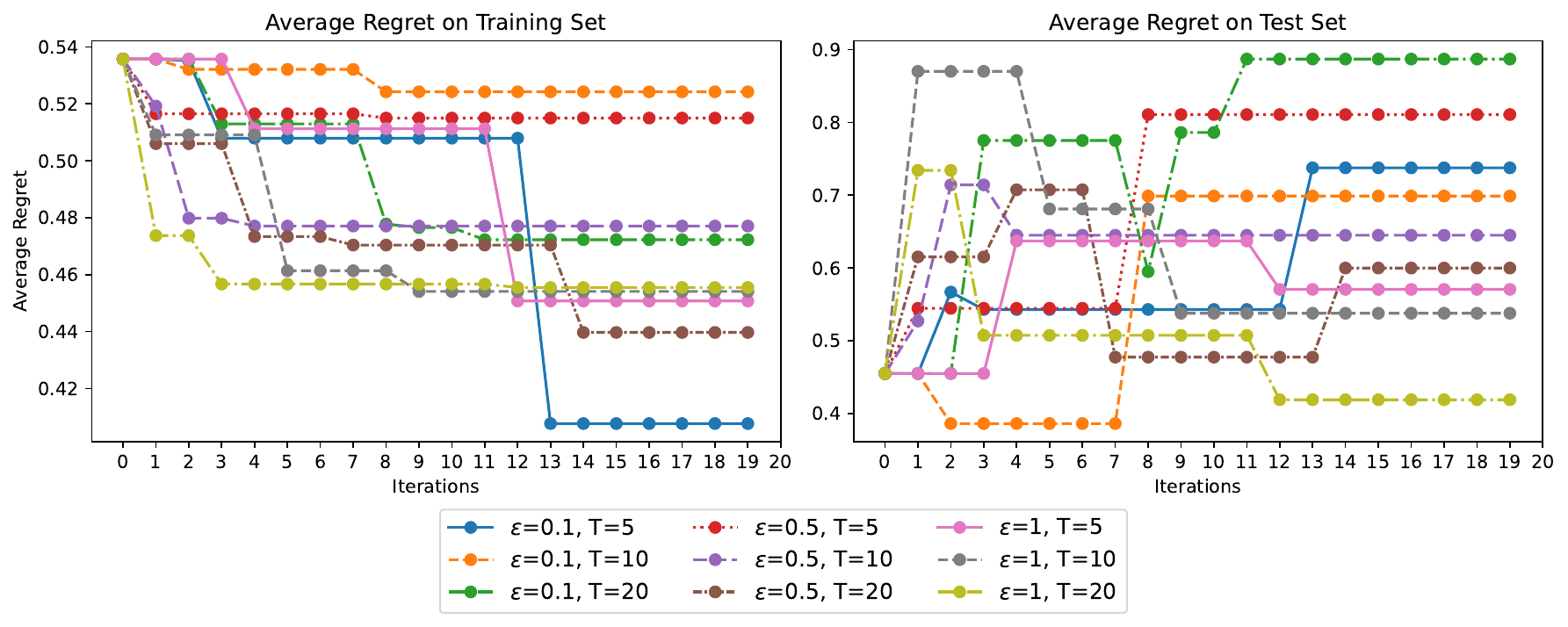}
	\caption{Execution of local search on a $10\times 13$ grid with different combinations of hyperparameters.}
	\label{exp_ls}
\end{figure}

In the implementation of Algorithm \ref{algorithm1}, we fixed the number of iterations to $20$. To fix the rest of the hyperparameters in this algorithm, we tested various combinations to choose the best. In Figure \ref{exp_ls}, we show the regrets on a training and test set instance of the bipartite matching problem, when running the local search algorithm starting from the SPO+ solution. We also ran similar tests for the shortest path instances.

Figure \ref{exp_ls} shows that $\epsilon = 1$ (size of the neighborhood) and $T=20$ (samples per iteration) provide the best trade-off between test and training performance; we thus set these values for experiments involving bipartite matching.
In the case of the shortest path, we determined that $\epsilon = 0.1$ and $T=20$ yields the best results.

In the case of the penalization method, to fix the value of $\kappa$, we also tested different variations ($\kappa\in \{0.1, 1, 10\}$) and observed that $\kappa = 0.1$ provided the best performance overall.\\

\paragraph{Hardware and Software}

We implemented all the aforementioned routines using Python 3.10. All non-convex quadratic models were solved using Gurobi 10.0.3 \citep{gurobi}. All experiments were run single-threaded on a Linux machine with an Intel Xeon Silver 4210 2.2G CPU and 128 GB RAM.

\paragraph{Repository} The codes and instances considered in this article can be found in the repository in the following \href{https://github.com/vbucarey/dfl_complexity_algorithms}{link}.

\subsection{Computational results}

In this section, we compare the decision-focused predictions obtained using the aforementioned methods by evaluating the regrets they achieve. To ensure the regrets are displayed on the same scale, and following the approach in \cite{elmachtoub2022smart}, we use the {\it normalized} regret instead of reporting the direct regret (as defined in \eqref{eq:mregret}). The normalized regret is defined as
\[\frac{ \mbox{Regret}(\mathcal{D}, \omega)}{\sum_{i \in [N]} z^*(c^i)}.\]
Henceforth, when we reference the regret, we mean this normalized version.

\subsubsection{Training set performance}
\begin{figure}
    \centering
    \includegraphics[width=1\linewidth]{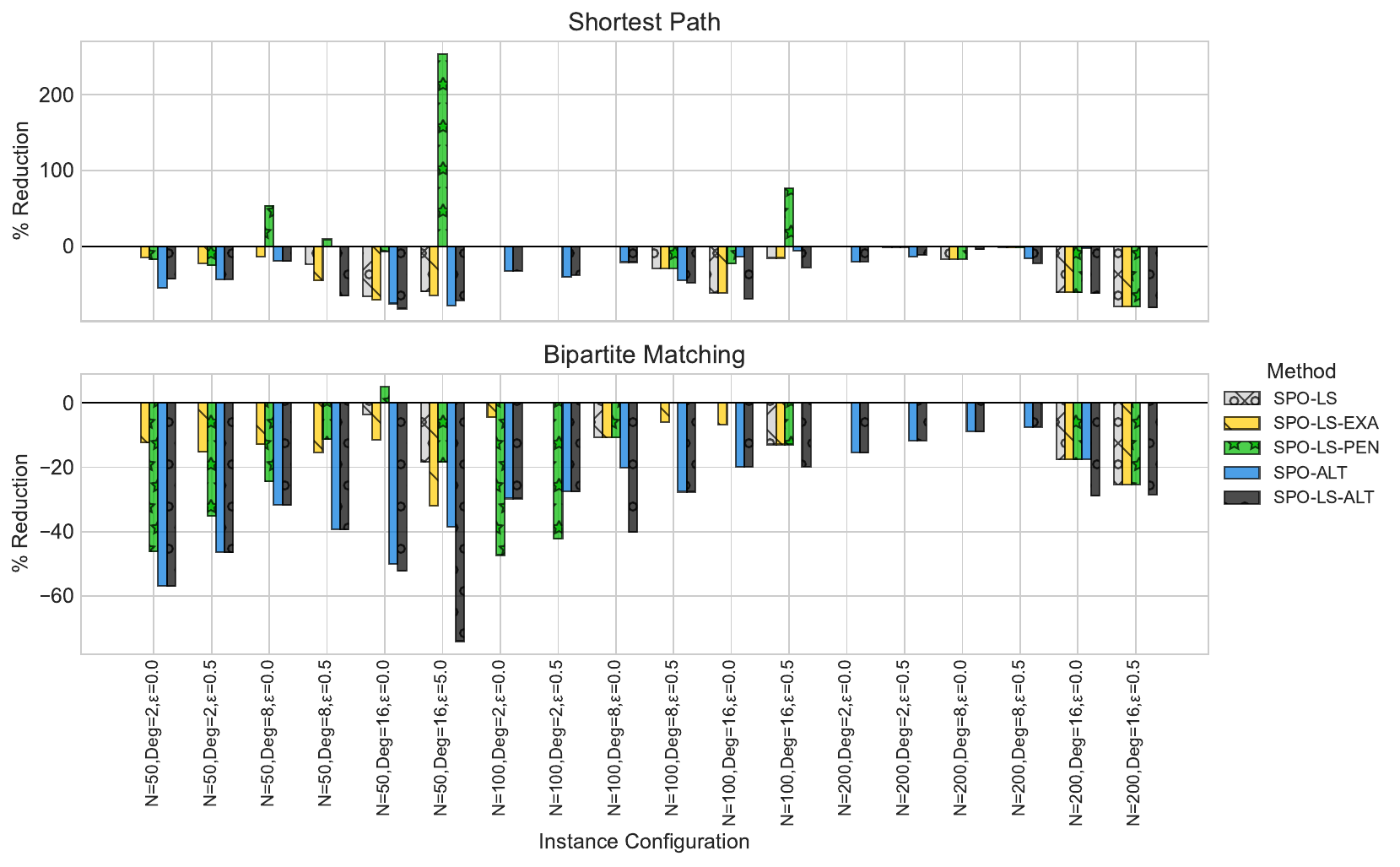}
    \caption{Training set performance on small shortest path (top) and bipartite matching (bottom) instances. Each bar represents the reduction/increment of normalized regret over the baseline SPO+.}
    \label{fig:train_performance}
\end{figure}

Figure \ref{fig:train_performance} summarizes the performance of the methods on the training set for small instances of shortest paths (top) and maximum weight bipartite matching (bottom). We use the regret returned by SPO \eqref{SPO+_lp} as the baseline, represented as the horizontal line at zero. Each bar displays the percentage decrease/increase in regret achieved by our methods. The method with the best performance, in terms of normalized regret, is the one with the most negative value.

In the shortest path instances, the alternating method, either by itself or in tandem with local search, achieves the best performance in terms of regret. We also observe that the penalized method may often improve (decrease) the regret, but in some instances, it can dramatically increase the regret (the bars with positive value). This occurs whenever the starting point is not valid for the slice given by the chosen penalty factor in \eqref{eq:penalized-final}, causing the solver to reject the solution. For bipartite matching, there are cases where the penalized method was the best. However, the alternating method obtained the overall best results.

We also note that, in some cases, the performance of the alternating method varies considerably depending on its starting point. In most cases, starting from the solution provided by the local search algorithm yields a better outcome than starting with SPO. 
However, overall, the results for the alternating method are fairly robust.

We believe that these results for the training set are highly encouraging. All methods we develop here are tailored for improving the training performance in moderate running times, to which we succeeded for these challenging instances. In Appendix C, we show the detailed values for Figure \ref{fig:train_performance}.\\

In what follows, we discuss two remaining computational aspects: test set performance and scalability.

\subsubsection{Test set performance}
Regarding the performance on the test set, the results are less conclusive: the performance of SPO can be improved, but there is no clear dominance of one method. 

\begin{figure}
    \centering
    \includegraphics[width=\linewidth]{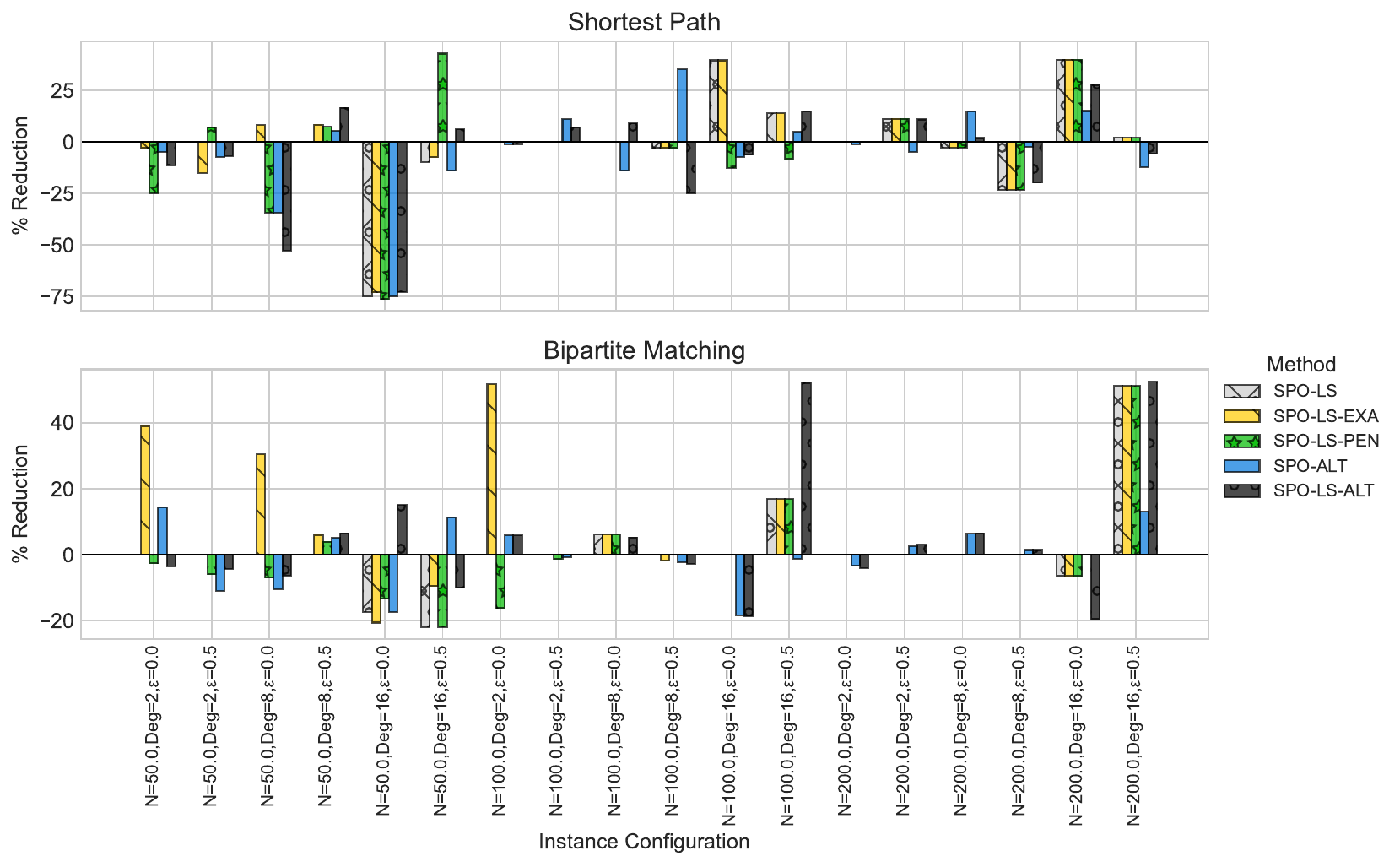}
    \caption{Test set performance on small shortest path (top) and bipartite matching (bottom) instances. Each bar represents the reduction/increment of normalized regret over the baseline SPO+.}
    \label{fig:test_results}
\end{figure}

In Figure \ref{fig:test_results} (top), we present results for shortest path instances. We observe that in 14 out of 16 cases, one of the methods we propose here improves the SPO performance. This means that at least one bar for each instance configuration is below 0. The two variants of the alternating method, which have the best in-training performance, have the best performance in 8 out of 18 test instances; however, they can increase the test regret in comparison to SPO in some cases. 
The results for bipartite matching, which we show in Figure \ref{fig:test_results} (bottom), are slightly different. Here, the alternating method, in its various implementations, achieves the best performance in 9 out of 18 instances --one more than in the shortest path instances. On the other hand, in 6 instances, no method was able to improve the test performance of SPO. In Appendix C, we show the detailed values for Figure \ref{fig:train_performance}.

Overall, from the test performance, we see the following takeaways: (1) SPO can provide a strong solution very quickly, (2) in many instances, the SPO performance can be improved considerably, and (3) which method yields the best improvement can vary significantly.

\subsubsection{Other performance metrics and scalability}

To better understand the performance of the different algorithms, in this section, we provide more details on their execution. 

Both the exact method (EXA) and the penalized method (PEN) often reach the time limit; in Table \ref{tab:exit_gap} we report the final gap values reported by Gurobi. The penalized method, in particular, shows a tendency to return exceedingly large optimality gaps. This occurs because PEN sometimes rejects the initial feasible solution and struggles to find either good bounds or high-quality feasible solutions. In one extreme case, the penalized method failed to identify a feasible solution for the shortest path instance entirely.
Also, note that the bipartite matching instances always finish with larger gaps than their shortest path counterpart.

These results indicate that, even though EXA provides an exact reformulation of \eqref{eq:bilevelpessimistic}, the current solver technology is still unable to provide a provably optimal solution in these instances.

\begin{table}[h!]
\centering
\small
\caption{Gap (percentage) returned by Gurobi at time limit for Exact and Penalized approaches.}
\label{tab:exit_gap}
%\resizebox{\textwidth}{!}{%
\begin{tabular}{cccrrrr} \toprule
\multicolumn{1}{l}{} & \multicolumn{1}{l}{} & \multicolumn{1}{l}{} & \multicolumn{2}{c}{\bf Shortest Path} & \multicolumn{2}{c}{\bf Bipartite Matching} \\ \midrule
\bf N &\bf Deg &\bf Noise &\bf SPO-LS-EXA &\bf SPO-LS-PEN &\bf SPO-LS-EXA &\bf SPO-LS-PEN \\ \midrule
50 & 2 & 0.0 & 3.4 & 2.8 & 6.0 & 3.7 \\
50 & 2 & 0.5 & 11.7 & 11.4 & 11.9 & 8.8 \\
50 & 8 & 0.0 & 10.4 & 17.3 & 42.9 & 52.4 \\
50 & 8 & 0.5 & 17.3 & 46.1 & 26.7 & 33.7 \\
50 & 16 & 0.0 & 27.0 & 10.9 & 43.8 & 155.6 \\
50 & 16 & 0.5 & 26.0 & 18.9 & 30.6 & 436.7 \\ \hline
100 & 2 & 0.0 & 7.2 & 8.5 & 10.0 & 5.0 \\
100 & 2 & 0.5 & 16.3 & 18.6 & 22.3 & 11.7 \\
100 & 8 & 0.0 & 28.5 & 42.9 & 53.6 & 138.1 \\
100 & 8 & 0.5 & 28.9 & 42.4 & 37.5 & 101.6 \\
100 & 16 & 0.0 & 55.7 & 43.4 & 70.8 & 930.8 \\
100 & 16 & 0.5 & 62.2 & 91.2 & 87.8 & 840.0 \\ \hline
200 & 2 & 0.0 & 8.2 & 9.3 & 11.5 & 14.0 \\
200 & 2 & 0.5 & 18.7 & 18.8 & 26.2 & 34.7 \\
200 & 8 & 0.0 & 36.6 & 45.9 & 56.4 & 121.8 \\
200 & 8 & 0.5 & 38.0 & 49.5 & 63.1 & 122.6 \\
200 & 16 & 0.0 & 45.8 & - & 65.0 & 1276.6 \\
200 & 16 & 0.5 & 56.0 & 58.6 & 78.6 & 1348.2 \\ \bottomrule
\end{tabular}%
%}
\end{table}

We also performed experiments on larger instances: with $N=1000$. Based on the analysis above, we exclude from further consideration the methods involving the exact method (EXA) and the penalized method (PEN), since their performance is dramatically affected by instance size.
We report our results in Tables \ref{tab:shortest1000} and \ref{tab:bimatching1000}. From these results, we see that our methods remain competitive, improving the performance of SPO in many cases. In two extreme cases, we obtained improvements of 25.9\% (for shortest path) and 39.1\% (for bipartite matching). However, overall, we note that these improvements start to become more modest than in the smaller instances. As before, the case of bipartite matching is harder to improve than the shortest path.\\

\begin{table}[htpb]
\centering
\small
\caption{Training and test performance on large ($N=1000$) shortest path instances}
\label{tab:shortest1000}
\resizebox{\textwidth}{!}{%
\begin{tabular}{ccrrrrrrrr}\toprule
 \multicolumn{1}{l}{} & \multicolumn{1}{l}{} & \multicolumn{4}{c}{\bf Train Set} & \multicolumn{4}{c}{\bf Test Set} \\ \midrule
\bf Deg &\bf Noise &\bf SPO &\bf SPO-LS &\bf SPO-LS-ALT &\bf SPO-ALT &\bf SPO &\bf SPO-LS &\bf SPO-LS-ALT &\bf SPO-ALT \\ \midrule
2 & 0.0 & 0.097 & 0.0\% & {\color[HTML]{F56B00} \textbf{-2.1\%}} & {\color[HTML]{F56B00} \textbf{-2.1\%}} & 0.1 & 0.0\% & {\color[HTML]{3531FF} \textbf{-2.0\%}} & {\color[HTML]{3531FF} \textbf{-2.0\%}} \\
 2 & 0.5 & 0.245 & 0.0\% & {\color[HTML]{F56B00} \textbf{-1.2\%}} & {\color[HTML]{F56B00} \textbf{-1.2\%}} & 0.268 & 0.0\% & -0.7\% & {\color[HTML]{3531FF} \textbf{-1.1\%}} \\
 8 & 0.0 & 0.532 & -6.4\% & {\color[HTML]{F56B00} \textbf{-9.0\%}} & -0.8\% & 0.601 & {\color[HTML]{3531FF} \textbf{-5.3\%}} & +0.7\% & -0.7\% \\
 8 & 0.5 & 0.678 & 0.0\% & {\color[HTML]{F56B00} \textbf{-4.4\%}} & {\color[HTML]{F56B00} \textbf{-4.4\%}} & 0.699 & 0.0\% & {\color[HTML]{3531FF} \textbf{-0.3\%}} & -0.1\% \\
 16 & 0.0 & 3.983 & -24.4\% & {\color[HTML]{F56B00} \textbf{-25.9\%}} & -0.5\% & 13.977 & -43.2\% & {\color[HTML]{3531FF} \textbf{-43.7\%}} & -0.0\% \\
 16 & 0.5 & 3.712 & -11.3\% & {\color[HTML]{F56B00} \textbf{-14.4\%}} & -0.2\% & 4.155 & {\color[HTML]{3531FF} \textbf{-2.0\%}} & -1.8\% & -0.4\% \\  \bottomrule
\end{tabular}%
}
\end{table}
\begin{table}[htpb]
\centering
\small
\caption{Training and test performance on large ($N=1000$) bipartite matching instances}
\label{tab:bimatching1000}
\resizebox{\textwidth}{!}{%
\begin{tabular}{ccrrrrrrrr}\toprule
\multicolumn{1}{l}{} & \multicolumn{1}{l}{} & \multicolumn{4}{c}{\bf Train Set} & \multicolumn{4}{c}{\bf Test Set} \\ \midrule
 \bf Deg &\bf Noise &\bf SPO &\bf SPO-LS &\bf SPO-LS-ALT &\bf SPO-ALT & \bf SPO &\bf SPO-LS &\bf SPO-LS-ALT &\bf SPO-ALT \\ \midrule
 2 & 0.0 & 0.117 & 0.0\% & {\color[HTML]{F56B00} \textbf{-1.7\%}} & {\color[HTML]{F56B00} \textbf{-1.7\%}} & {\color[HTML]{3531FF} \textbf{0.118}} & 0.0\% & 0.0\% & 0.0\% \\
 2 & 0.5 & 0.225 & 0.0\% & {\color[HTML]{F56B00} \textbf{-0.9\%}} & {\color[HTML]{F56B00} \textbf{-0.9\%}} & {\color[HTML]{3531FF} \textbf{0.242}} & 0.0\% & +0.4\% & +0.4\% \\
 8 & 0.0 & 0.406 & 0.0\% & {\color[HTML]{F56B00} \textbf{-0.7\%}} & {\color[HTML]{F56B00} \textbf{-0.7\%}} & 0.416 & 0.0\% & {\color[HTML]{3531FF} \textbf{-0.5\%}} & {\color[HTML]{3531FF} \textbf{-0.5\%}} \\
 8 & 0.5 & 0.411 & 0.0\% & {\color[HTML]{F56B00} \textbf{-1.5\%}} & {\color[HTML]{F56B00} \textbf{-1.5\%}} & {\color[HTML]{3531FF} \textbf{0.449}} & 0.0\% & 0.0\% & 0.0\% \\
 16 & 0.0 & 0.673 & -11.0\% & {\color[HTML]{F56B00} \textbf{ -11.6\%}} & -0.9\% & {\color[HTML]{3531FF} \textbf{0.652}} & +6.3\% & +6.4\% & +0.5\% \\
 16 & 0.5 & 0.604 & -38.9\% & {\color[HTML]{F56B00} \textbf{-39.1\%}} & -0.3\% & 0.697 & -4.9\% & {\color[HTML]{3531FF} \textbf{-5.0\%}} & +1.0\% \\ \bottomrule
\end{tabular}%
}
\end{table}

Finally, to provide a more fleshed-out analysis on the alternating method and shed light on the reduced improvements for large instances, we show per-iteration statistics in Table \ref{tab:performance_metrics}. This table shows the number of iterations and average time-per-iteration of the two versions of the alternating method: starting from SPO directly or from local search.

\begin{table}[h!]
\centering
\small
\caption{Detailed performance metrics of alternating method. The reported times are the average iteration time in the alternating method.}
\label{tab:performance_metrics}
%\resizebox{\textwidth}{!}{%
\begin{tabular}{cccrrrr}
\toprule
 & & & \multicolumn{2}{c}{\textbf{Shortest Path}} & \multicolumn{2}{c}{\textbf{Bipartite Matching}} \\
\midrule
\textbf{N} & \textbf{Deg} & \textbf{Noise} & \textbf{SPO-LS-ALT} & \textbf{SPO-ALT} & \textbf{SPO-LS-ALT} & \textbf{SPO-ALT} \\
 & & & Iterations (Time) & Iterations (Time) & Iterations (Time) & Iterations (Time) \\
\midrule
50  & 2  & 0.0 & 332 (1.60s) & 456 (1.62s) & 1922 (0.83s) & 2308 (0.87s) \\
50  & 2  & 0.5 & 327 (1.63s) & 471 (1.65s) & 2197 (0.72s) & 2262 (0.72s) \\
50  & 8  & 0.0 & 353 (1.63s) & 448 (1.65s) & 1717 (1.22s) & 1759 (1.23s) \\
50  & 8  & 0.5 & 610 (1.09s) & 583 (1.11s) & 1770 (1.14s) & 1812 (1.14s) \\
50  & 16 & 0.0 & 318 (1.62s) & 435 (1.95s) & 1792 (1.28s) & 1688 (1.20s) \\
50  & 16 & 0.5 & 333 (1.67s) & 449 (1.61s) & 1379 (1.24s) & 1754 (1.59s) \\
\midrule
100 & 2  & 0.0 & 141 (6.15s) & 182 (5.86s) & 862 (2.39s) & 872 (2.33s) \\
100 & 2  & 0.5 & 114 (6.99s) & 198 (7.13s) & 880 (2.41s) & 852 (2.33s) \\
100 & 8  & 0.0 & 115 (6.59s) & 202 (7.01s) & 918 (2.25s) & 896 (2.16s) \\
100 & 8  & 0.5 & 121 (7.09s) & 177 (5.94s) & 754 (2.85s) & 763 (2.79s) \\
100 & 16 & 0.0 & 182 (4.46s) & 208 (4.70s) & 741 (2.88s) & 802 (2.90s) \\
100 & 16 & 0.5 & 160 (4.28s) & 223 (4.55s) & 830 (2.69s) & 796 (2.57s) \\
\midrule
200 & 2  & 0.0 & 50 (24.01s) & 73 (22.85s) & 298 (6.34s) & 397 (6.53s) \\
200 & 2  & 0.5 & 50 (16.73s) & 84 (21.36s) & 266 (7.44s) & 348 (7.59s) \\
200 & 8  & 0.0 & 52 (20.27s) & 77 (19.27s) & 310 (6.89s) & 342 (6.82s) \\
200 & 8  & 0.5 & 49 (18.96s) & 78 (21.53s) & 259 (8.07s) & 320 (8.09s) \\
200 & 16 & 0.0 & 59 (13.48s) & 116 (14.69s) & 326 (6.66s) & 349 (6.36s) \\
200 & 16 & 0.5 & 53 (15.19s) & 83 (19.31s) & 318 (6.69s) & 341 (6.56s) \\
\midrule
1000 & 2  & 0.0 & 32 (10.36s) & 49 (10.41s) & 173 (5.42s) & 256 (5.45s) \\
1000 & 2  & 0.5 & 33 (7.61s)  & 50 (7.03s)  & 184 (4.39s) & 276 (4.42s) \\
1000 & 8  & 0.0 & 33 (7.74s)  & 50 (7.59s)  & 167 (5.76s) & 250 (5.78s) \\
1000 & 8  & 0.5 & 33 (7.88s)  & 50 (7.82s)  & 166 (6.08s) & 245 (5.86s) \\
1000 & 16 & 0.0 & 33 (8.29s)  & 50 (8.47s)  & 151 (6.70s) & 233 (7.32s) \\
1000 & 16 & 0.5 & 33 (8.90s)  & 49 (7.76s)  & 146 (6.61s) & 230 (7.69s) \\
\bottomrule
\end{tabular}
%}
\end{table}

From Table \ref{tab:performance_metrics}, we note that the number of iterations reduces dramatically as $N$ increases: this can partially explain why we observed more moderate improvements with respect to SPO in $N=1000$.
We also note that, even if running local search reduces the number of iterations that ALT can perform (since the budget time is shared), the results in Tables \ref{tab:shortest1000} and \ref{tab:bimatching1000} suggest that it may be worth running them in tandem. 

We believe that these results suggest that a batch version of the alternating method can be worthwhile for large instances. We strongly believe that this, along with other computational enhancements, can scale the strong result we observed in small instances.

\section{Conclusions} \label{sec:con}
	
In this work, we present an in-depth analysis of the optimization problem behind the training task of decision-focused learning.
Our proof of membership in NP indicates that this problem is not higher in the computational complexity hierarchy, unless the latter collapses. In addition, we show that the problem of determining if regret zero is achievable or not is polynomial-time solvable under mild assumptions.
Additionally, we derive a non-convex quadratic optimization reformulation of the problem, whose structure we exploit empirically. Furthermore, by leveraging intermediate steps of the reformulation, we develop algorithms to address the problem, including a local search procedure and an alternating direction method. These two algorithms only require solving linear programs at each iteration. When performed in tandem, they can provide predictions with strong performance on both training and test sets for challenging shortest path instances and maximum weight matching problems with unknown cost/weight vectors.

Our results show that improvements (both in training and test instances) over SPO+ --a state-of-the-art method-- can be achieved, thus effectively producing better decision-focused predictive models. The main drawback we observe is scalability: when the number of observations is large, the methods we show here, while still competitive, provide more moderate improvements. However, our results show great potential of our non-convex optimization framework, and we strongly believe that after computational enhancements, such as a batch version of the alternating algorithm, these methods can achieve large-scale tractability.

\bibliographystyle{abbrv} % outcomment this and next line in Case 1
\bibliography{biblio} % if more than one, comma separated
% Acknowledgments here
\paragraph{Acknowledgments}{The authors would like to thank Paul Grigas for helpful comments on this work.}

\newpage

\begin{appendix}

\section*{Appendix A: Proof of Corollary 3}
\label{proof_corollary3}

\noindent{\it Corollary 3:} Consider \eqref{expected_regret_sample} when $m(\omega, x)= \omega x$. The regret function \eqref{eq:mregret} only has a finite number of values, and, furthermore, the minimum regret \eqref{expected_regret_sample} is always attained.

\begin{proof}
Let $I^i$ be an arbitrary set of indices of active constraints defining a face $F^i$ of $V$ for the $i$-th follower. We can consider the following system, which is similar to \eqref{eq:system}: 
\begin{subequations}\label{eq:system2}
\begin{align}
    (\rho^i)^\top A & =  \omega x^i  && \forall i\in [N]\\
    \rho^i_j & >  0 && \forall\, j\in I^i,\, \forall i\in [N]\\
    \rho^i_j &= 0 && \forall\, j\not\in I^i,\, \forall i\in [N]
\end{align}
\end{subequations}
If \eqref{eq:system2} is infeasible, there is no $\omega$ ``consistent'' with those faces. And if \eqref{eq:system2} is feasible, every $\omega$ that is valid for \eqref{eq:system2} satisfies $V^*(\omega x^i) = F^i$, thus the regret is the same for all of them. Since the number of possible $(I^i)_{i\in [N]}$ is finite, we conclude.
\end{proof}

\section*{Appendix B: Proof of Theorem 2}

\label{theoremZeroRegret}

\noindent {\bf Theorem 2:} If the input for $\mbox{SIMPLE-REGRET}$ is restricted to $M=0$ (i.e., determining if there is a solution with zero regret) and the data $(c^i, x^i)_{i=1}^N$ and the polytope $V$ satisfy Assumption \ref{assumption:uniqueopt}, then the problem can be solved in polynomial time.

\begin{proof}
This proof uses similar concepts to the proof of Theorem \ref{theo:membershipNP}, but there are some important differences.
We begin by noting that, since $v^i \in V$, it always holds that $c^i{}^\top v^i - z^*(c^i) \geq 0$, i.e., the regret is always nonnegative. Thus, for $M=0$, $\mbox{SIMPLE-REGRET}$ outputs `Yes' if and only if there exists $\omega$ such that
    \begin{equation}\label{eq:zeroregret-unique}
    \max_{v^i \in V^*(\omega x^i  )} c^i{}^\top v^i - z^*(c^i) = 0 \quad \forall i\in [N]
    \end{equation}
    In other words, the optimal face of 
    \[\max_{v^i \in V^*(\omega x^i  )} c^i{}^\top v^i\]
    is contained in the optimal face defining $z^*(c^i)$, i.e., $V^*(c^i)$. 
    By Assumption \ref{assumption:uniqueopt}, $V^*(c^i)$ is a singleton.
    
    Let $I^i$ be the indices of \emph{every} active constraint at $V^*(c^i)$ (which can be computed in polynomial time). 
    We claim that \eqref{eq:zeroregret-unique} holds for some $\tilde{\omega}$ if and only if the following \emph{linear} system (over variables $\rho, \omega$) is feasible:
    \begin{subequations}\label{eq:thesystem-unique}
    \begin{align}
        %A v^i & \leq b \\
        %a_j^\top v^i - b_j &= 0 && \forall j\in I^i,\, \forall i\in [N]\\
        (\rho^i)^\top A &= x^i \omega && \forall i \in [N]\\
        %\rho^i_j &\leq -1 && \forall j,\, \hat{\rho}^i_j < 0,\, \forall i\in [N]\\
        \rho^i_j &= 0 && \forall j\not\in I^i,\, \forall i\in [N] \\
        \rho^i_j & \geq 1 && \forall j\in I^i,\, \forall i\in [N] 
    \end{align}
    \end{subequations}

    We note that the $\tilde{\omega}$ certifying \eqref{eq:zeroregret-unique} may or may not be the same as the $\omega$ in \eqref{eq:thesystem-unique}.
    Proving this equivalence suffices, as system \eqref{eq:thesystem-unique} can be solved in polynomial time.\\
    
    Suppose the system \eqref{eq:thesystem-unique} is feasible, and take $(\rho, \omega)$ that satisfy it. 
    Additionally, for each $i$, take $v^i$ the unique optimal solution in $V^*(c^i)$.
    We claim that each $(v^i, \rho^i)$ optimize $V^*(\omega x^i)$ and its dual. Indeed, primal and dual feasibility hold by construction. Complementary slackness also holds by construction, as for each $j$ either $a_j^\top v^i - b_j=0$ or $\rho^i_j = 0$.

    Fixing the dual solutions $\rho^i$, and via complementary slackness again, we see that \emph{any} $\tilde{v}^i\in V^*(\omega x^i)$, must satisfy $a_j^\top \tilde{v}^i - b_j = 0$  $\forall j \in I^i$. This means that $V^*(\omega x^i)$ is a singleton. This implies \eqref{eq:zeroregret-unique} using $\tilde{\omega} = \omega$.\\

    For the other direction, suppose \eqref{eq:zeroregret-unique} holds for some $\omega$.
    This directly implies that $V^*(\omega x^i) = V^*(c^i) = \{v^i\}$, since \emph{any} other vector in the polytope has a strictly larger value than $c^i{}^\top v^i$ by assumption. 
    By optimality conditions for $V^*(\omega x^i)$, there exists $\tilde{\rho}$ such that
    \begin{align*}
        (\tilde{\rho}^i)^\top A &= x^i \omega && \forall i \in [N]\\
        \tilde{\rho}^i & \geq 0 && \forall i \in [N]\\
        \tilde{\rho}^i_j (a_j^\top v^i - b) &= 0 && \forall j,\, \forall i\in [N] 
    \end{align*}

    Furthermore, we can take $\tilde{\rho}^i$ to be strictly complementary with $v^i$ (as the latter is the unique optimal solution). Thus, the following holds
    \begin{align*}
        (\tilde{\rho}^i)^\top A &= x^i \omega && \forall i \in [N]\\
        \tilde{\rho}^i_j & > 0 && \forall j\in I^i,\, \forall i \in [N]\\ 
        \tilde{\rho}^i_j & = 0 && \forall j\not\in I^i,\, \forall i \in [N]
    \end{align*}

    We can then rescale $\tilde{\rho}$ and $\omega$ to obtain \eqref{eq:thesystem-unique}.
\end{proof}

\newpage
\section*{Appendix C: Numerical results of Figures 3 and 4} \label{appendix:training_results}

In Tables \ref{tab:sp_train_small} and \ref{tab:bm_train_small} for training and \ref{tab:sp_test_small} and \ref{tab:bm_test_small} for test, we present the detailed results of Figures 3 and 4, respectively. We use the regret returned by SPO (model \eqref{SPO+_lp}) as the baseline. The subsequent columns display the percentage decrease in regret achieved by our methods. The method with the best performance in terms of normalized regret is highlighted in color.
 
\begin{table}[htpb]
\small
\caption{Training set performance on small shortest path instances. The first three columns specify the instance parameters. The column labeled SPO provides the normalized regret achieved by the SPO+ method. Subsequent columns provide the improvements over SPO. Best regrets are highlighted in color.}
\label{tab:sp_train_small}
\resizebox{\textwidth}{!}{%
\begin{tabular}{cccrrrrrr}
\toprule
N & Deg & Noise & SPO & SPO-LS & SPO-LS-EXA & SPO-LS-PEN & SPO-LS-ALT & SPO-ALT \\ \midrule
50 & 2 & 0.0 & 0.035 & 0.0\% & -14.3\% & -17.1\% & -42.9\% & {\color[HTML]{F56B00} \textbf{-54.3\%}} \\
50 & 2 & 0.5 & 0.171 & 0.0\% & -22.2\% & -25.1\% & {\color[HTML]{F56B00} \textbf{-43.3\%}} & {\color[HTML]{F56B00} \textbf{-43.3\%}} \\
50 & 8 & 0.0 & 0.133 & 0.0\% & -13.5\% & \textbf{+53.4\%} & {\color[HTML]{F56B00} \textbf{-19.5\%}} & {\color[HTML]{F56B00} \textbf{-19.5\%}} \\
50 & 8 & 0.5 & 0.381 & -23.6\% & -45.1\% & \textbf{+9.2\%} & {\color[HTML]{F56B00} \textbf{-67.5\%}} & -64.8\% \\
50 & 16 & 0.0 & 1.254 & -66.4\% & -70.6\% & -6.7\% & {\color[HTML]{F56B00} \textbf{-82.0\%}} & -75.4\% \\
50 & 16 & 0.5 & 0.99 & -59.5\% & -64.5\% & \textbf{+253.5\%} & -71.2\% & {\color[HTML]{F56B00} \textbf{-77.8\%}} \\ \hline
100 & 2 & 0.0 & 0.078 & 0.0\% & 0.0\% & 0.0\% & {\color[HTML]{F56B00} \textbf{-32.1\%}} & {\color[HTML]{F56B00} \textbf{-32.1\%}} \\
100 & 2 & 0.5 & 0.195 & 0.0\% & 0.0\% & 0.0\% & -37.9\% & {\color[HTML]{F56B00} \textbf{-40.5\%}} \\
100 & 8 & 0.0 & 0.399 & 0.0\% & 0.0\% & 0.0\% & {\color[HTML]{F56B00} \textbf{-20.8\%}} & {\color[HTML]{F56B00} \textbf{-20.8\%}} \\
100 & 8 & 0.5 & 0.572 & -29.0\% & -29.0\% & -29.0\% & {\color[HTML]{F56B00} \textbf{-47.6\%}} & -44.9\% \\
100 & 16 & 0.0 & 3.243 & -61.2\% & -61.2\% & -23.0\% & {\color[HTML]{F56B00} \textbf{-69.1\%}} & -13.8\% \\
100 & 16 & 0.5 & 1.945 & -15.3\% & -15.3\% & \textbf{+76.1\%} & {\color[HTML]{F56B00} \textbf{-28.0\%}} & -5.3\% \\ \hline
200 & 2 & 0.0 & 0.089 & 0.0\% & 0.0\% & 0.0\% & {\color[HTML]{F56B00} \textbf{-20.2\%}} & {\color[HTML]{F56B00} \textbf{-20.2\%}} \\
200 & 2 & 0.5 & 0.232 & -0.9\% & -0.9\% & -0.9\% & -11.2\% & {\color[HTML]{F56B00} \textbf{-13.4\%}} \\
200 & 8 & 0.0 & 0.693 & -16.7\% & -16.7\% & -16.7\% & {\color[HTML]{F56B00} \textbf{-25.4\%}} & -4.0\% \\
200 & 8 & 0.5 & 0.621 & -1.4\% & -1.4\% & -1.4\% & {\color[HTML]{F56B00} \textbf{-22.1\%}} & -16.1\% \\
200 & 16 & 0.0 & 2.14 & -60.5\% & -60.5\% & -60.5\% & {\color[HTML]{F56B00} \textbf{-61.5\%}} & -2.9\% \\
200 & 16 & 0.5 & 6.194 & -79.4\% & -79.4\% & -79.4\% & {\color[HTML]{F56B00} \textbf{-80.4\%}} & -0.2\% \\ \bottomrule
\end{tabular}%
}
\end{table}
\begin{table}[h]
\small
\caption{Training set performance on small bipartite matching instances. The first three columns specify the instance parameters. The column labeled SPO provides the normalized regret achieved by the SPO+ method. Subsequent columns provide the improvements over SPO. Best regrets are highlighted in color.}
\label{tab:bm_train_small}
\resizebox{\textwidth}{!}{%
\begin{tabular}{cccrrrrrr}
\toprule
N & Deg & Noise & SPO & SPO-LS & SPO-LS-EXA & SPO-LS-PEN & SPO-LS-ALT & SPO-ALT \\ \midrule
50 & 2 & 0.0 & 0.065 & 0.0\% & -12.3\% & -46.2\% & {\color[HTML]{F56B00} \textbf{-56.9\%}} & {\color[HTML]{F56B00} \textbf{-56.9\%}} \\
50 & 2 & 0.5 & 0.125 & 0.0\% & -15.2\% & -35.2\% & {\color[HTML]{F56B00} \textbf{-46.4\%}} & {\color[HTML]{F56B00} \textbf{-46.4\%}} \\
50 & 8 & 0.0 & 0.344 & 0.0\% & -12.8\% & -24.4\% & {\color[HTML]{F56B00} \textbf{-31.7\%}} & {\color[HTML]{F56B00} \textbf{-31.7\%}} \\
50 & 8 & 0.5 & 0.249 & 0.0\% & -15.3\% & -11.2\% & {\color[HTML]{F56B00} \textbf{-39.4\%}} & {\color[HTML]{F56B00} \textbf{-39.4\%}} \\
50 & 16 & 0.0 & 0.316 & -3.5\% & -11.4\% & \textbf{+5.1\%} & {\color[HTML]{F56B00} \textbf{-52.2\%}} & -50.0\% \\
50 & 16 & 0.5 & 0.344 & -18.3\% & -32.0\% & -18.3\% & {\color[HTML]{F56B00} \textbf{-74.1\%}} & -38.4\% \\ \hline
100 & 2 & 0.0 & 0.091 & 0.0\% & -4.4\% & {\color[HTML]{F56B00} \textbf{-47.3\%}} & -29.7\% & -29.7\% \\
100 & 2 & 0.5 & 0.182 & 0.0\% & 0.0\% & {\color[HTML]{F56B00} \textbf{-42.3\%}} & -27.5\% & -27.5\% \\
100 & 8 & 0.0 & 0.391 & -10.7\% & -10.7\% & -10.7\% & {\color[HTML]{F56B00} \textbf{-40.2\%}} & -20.2\% \\
100 & 8 & 0.5 & 0.29 & 0.0\% & -5.9\% & 0.0\% & {\color[HTML]{F56B00} \textbf{-27.6\%}} & {\color[HTML]{F56B00} \textbf{-27.6\%}} \\
100 & 16 & 0.0 & 0.415 & 0.0\% & -6.7\% & 0.0\% & {\color[HTML]{F56B00} \textbf{-19.8\%}} & {\color[HTML]{F56B00} \textbf{-19.8\%}} \\
100 & 16 & 0.5 & 0.536 & -12.9\% & -12.9\% & -12.9\% & {\color[HTML]{F56B00} \textbf{-38.6\%}} & -19.8\% \\ \hline
200 & 2 & 0.0 & 0.103 & 0.0\% & 0.0\% & 0.0\% & {\color[HTML]{F56B00} \textbf{-15.5\%}} & {\color[HTML]{F56B00} \textbf{-15.5\%}} \\
200 & 2 & 0.5 & 0.207 & 0.0\% & 0.0\% & 0.0\% & {\color[HTML]{F56B00} \textbf{-11.6\%}} & {\color[HTML]{F56B00} \textbf{-11.6\%}} \\
200 & 8 & 0.0 & 0.361 & 0.0\% & 0.0\% & 0.0\% & {\color[HTML]{F56B00} \textbf{-8.9\%}} & {\color[HTML]{F56B00} \textbf{-8.9\%}} \\
200 & 8 & 0.5 & 0.387 & 0.0\% & 0.0\% & 0.0\% & {\color[HTML]{F56B00} \textbf{-7.5\%}} & {\color[HTML]{F56B00} \textbf{-7.5\%}} \\
200 & 16 & 0.0 & 0.478 & -17.6\% & -17.6\% & -17.6\% & {\color[HTML]{F56B00} \textbf{-28.9\%}} & -17.6\% \\
200 & 16 & 0.5 & 0.589 & -25.3\% & -25.3\% & -25.3\% & {\color[HTML]{F56B00} \textbf{-33.8\%}} & -28.5\% \\ \bottomrule
\end{tabular}%
}
\end{table}
\begin{table}[h!]
\centering
\small
\caption{Test set performance on small shortest path instances. The first three columns specify the instance parameters. The column labeled SPO provides the normalized regret achieved by the SPO+ method. Subsequent columns provide the improvements over SPO. Best regrets are highlighted in color.}
\label{tab:sp_test_small}
\resizebox{\textwidth}{!}{%
\begin{tabular}{cccrrrrrr}
\toprule
\bf N & \bf Deg &\bf Noise &\bf SPO &\bf SPO-LS &\bf SPO-LS-EXA &\bf SPO-LS-PEN &\bf SPO-LS-ALT & \bf SPO-ALT \\ \midrule
50 & 2 & 0.0 & 0.105 & 0.0\% & -2.9\% & {\color[HTML]{3531FF} \textbf{-24.8\%}} & -11.4\% & -4.8\% \\
50 & 2 & 0.5 & 0.304 & 0.0\% & {\color[HTML]{3531FF} \textbf{-15.1\%}} & +6.9\% & -6.9\% & -7.2\% \\
50 & 8 & 0.0 & 0.45 & 0.0\% & +8.2\% & -34.2\% & {\color[HTML]{3531FF} \textbf{-52.7\%}} & -34.4\% \\
50 & 8 & 0.5 & 1.056 & {\color[HTML]{3531FF} \textbf{-3.3\%}} & +8.4\% & +7.4\% & +16.5\% & +5.2\% \\
50 & 16 & 0.0 & 3.762 & -74.9\% & -73.1\% & {\color[HTML]{3531FF} \textbf{-76.2\%}} & -72.8\% & -74.9\% \\
50 & 16 & 0.5 & 3.504 & -9.6\% & -7.4\% & +42.9\% & +6.2\% & {\color[HTML]{3531FF} \textbf{-13.8\%}} \\ \hline
100 & 2 & 0.0 & 0.099 & 0.0\% & 0.0\% & 0.0\% & {\color[HTML]{3531FF} \textbf{-1.0\%}} & {\color[HTML]{3531FF} \textbf{-1.0\%}} \\
100 & 2 & 0.5 & {\color[HTML]{3531FF} \textbf{0.227}} & 0.0\% & 0.0\% & 0.0\% & +7.0\% & +11.0\% \\
100 & 8 & 0.0 & 0.776 & 0.0\% & 0.0\% & 0.0\% & +8.9\% & {\color[HTML]{3531FF} \textbf{-13.9\%}} \\
100 & 8 & 0.5 & 0.707 & -3.0\% & -3.0\% & -3.0\% & {\color[HTML]{3531FF} \textbf{-24.9\%}} & +35.5\% \\
100 & 16 & 0.0 & 3.882 & +39.6\% & +39.6\% & {\color[HTML]{3531FF} \textbf{-12.6\%}} & -6.1\% & -7.5\% \\
100 & 16 & 0.5 & 13.645 & +13.9\% & +13.9\% & {\color[HTML]{3531FF} \textbf{-8.2\%}} & +14.9\% & +5.0\% \\ \hline
200 & 2 & 0.0 & 0.099 & 0.0\% & 0.0\% & 0.0\% & 0.0\% & {\color[HTML]{3531FF} \textbf{-1.0\%}} \\ 
200 & 2 & 0.5 & 0.238 & +11.3\% & +11.3\% & +11.3\% & +10.9\% & {\color[HTML]{3531FF} \textbf{-5.0\%}} \\
200 & 8 & 0.0 & 0.591 & {\color[HTML]{3531FF} \textbf{-2.7\%}} & {\color[HTML]{3531FF} \textbf{-2.7\%}} & {\color[HTML]{3531FF} \textbf{-2.7\%}} & +1.9\% & +14.7\% \\
200 & 8 & 0.5 & 1.023 & {\color[HTML]{3531FF} \textbf{-23.3\%}} & {\color[HTML]{3531FF} \textbf{-23.3\%}} & {\color[HTML]{3531FF} \textbf{-23.3\%}} & -19.8\% & -2.5\% \\
200 & 16 & 0.0 & {\color[HTML]{3531FF} \textbf{1.836}} & +39.9\% & +39.9\% & +39.9\% & +27.6\% & +15.0\% \\
200 & 16 & 0.5 & 3.384 & +2.0\% & +2.0\% & +2.0\% & -5.8\% & {\color[HTML]{3531FF} \textbf{-12.1\%}} \\ \bottomrule
\end{tabular}%
}
\end{table}
\begin{table}[h]
\small
\caption{Test set performance on small bipartite matching instances. The first three columns specify the instance parameters. The column labeled SPO provides the normalized regret achieved by the SPO+ method. Subsequent columns provide the improvements over SPO. Best regrets are highlighted in color.}
\label{tab:bm_test_small}
\resizebox{\textwidth}{!}{%
\begin{tabular}{cccrrrrrr}
\toprule
\bf N  &\bf Deg &\bf Noise &\bf SPO &\bf SPO-LS &\bf SPO-LS-EXA &\bf SPO-LS-PEN &\bf SPO-LS-ALT &\bf SPO-ALT \\ \midrule
50 & 2 & 0.0 & 0.118 & 0.0\% & +39.0\% & -2.5\% & {\color[HTML]{3531FF} \textbf{-3.4\%}} & +14.4\% \\
50 & 2 & 0.5 & 0.26 & 0.0\% & 0.0\% & -5.8\% & -4.2\% & {\color[HTML]{3531FF} \textbf{-10.8\%}} \\
50 & 8 & 0.0 & 0.349 & 0.0\% & +30.4\% & -6.9\% & -6.3\% & {\color[HTML]{3531FF} \textbf{-10.3\%}} \\
50 & 8 & 0.5 & {\color[HTML]{3531FF} \textbf{0.474}} & 0.0\% & +6.1\% & +4.0\% & +6.5\% & +5.3\% \\
50 & 16 & 0.0 & 0.781 & -17.2\% & -20.5\% & -13.3\% & +15.1\% & {\color[HTML]{3531FF} \textbf{-17.4\%}} \\
50 & 16 & 0.5 & 0.596 & {\color[HTML]{3531FF} \textbf{-21.8\%}} & -9.4\% & {\color[HTML]{3531FF} \textbf{-21.8\%}} & -9.9\% & +11.4\% \\ \hline
100 & 2 & 0.0 & 0.118 & 0.0\% & +51.7\% & {\color[HTML]{3531FF} \textbf{-16.1\%}} & +5.9\% & +5.9\% \\
100 & 2 & 0.5 & 0.252 & 0.0\% & 0.0\% & {\color[HTML]{3531FF} \textbf{-1.2\%}} & 0.0\% & -0.8\% \\
100 & 8 & 0.0 & {\color[HTML]{3531FF} \textbf{0.403}} & +6.2\% & +6.2\% & +6.2\% & +5.2\% & 0.0\% \\
100 & 8 & 0.5 & 0.431 & 0.0\% & -1.6\% & 0.0\% & {\color[HTML]{3531FF} \textbf{-2.6\%}} & -2.1\% \\
100 & 16 & 0.0 & 0.734 & 0.0\% & 0.1\% & 0.0\% & {\color[HTML]{3531FF} \textbf{-18.5\%}} & -18.4\% \\
100 & 16 & 0.5 & 0.455 & +16.9\% & +16.9\% & +16.9\% & +52.1\% & {\color[HTML]{3531FF} \textbf{-1.3\%}} \\ \hline
200 & 2 & 0.0 & 0.126 & 0.0\% & 0.0\% & 0.0\% & {\color[HTML]{3531FF} \textbf{-4.0\%}} & -3.2\% \\
200 & 2 & 0.5 & {\color[HTML]{3531FF} \textbf{0.229}} & 0.0\% & 0.0\% & 0.0\% & +3.1\% & +2.6\% \\
200 & 8 & 0.0 & {\color[HTML]{3531FF} \textbf{0.368}} & 0.0\% & 0.0\% & 0.0\% & +6.5\% & +6.5\% \\
200 & 8 & 0.5 & {\color[HTML]{3531FF} \textbf{0.401}} & 0.0\% & 0.0\% & 0.0\% & +1.5\% & +1.5\% \\
200 & 16 & 0.0 & 0.739 & -6.4\% & -6.4\% & -6.4\% & {\color[HTML]{3531FF} \textbf{-19.4\%}} & +0.1\% \\
200 & 16 & 0.5 & {\color[HTML]{3531FF} \textbf{0.455}} & +51.2\% & +51.2\% & +51.2\% & +52.5\% & +13.2\% \\ \bottomrule
\end{tabular}%
}
\end{table}

\end{appendix}
\end{document}